# Delegating Earth science model calibration to knowledge-guided LLM agents

Zhanliang Zhu[1,2], Ziwei Li[1,2,*], Liujun Zhu[1,2], Yuchen Liu[3], Junliang Jin[1,2,4,*], Jianyun Zhang[4]

[1]State Key Laboratory of Water Disaster Prevention, Hohai University, Nanjing 210098, China.
[2]Yangtze Institute for Conservation and Development, Hohai University, Nanjing 210098, China.
[3]Meta Platforms Inc., Menlo Park, California, USA.
[4]Nanjing Hydraulic Research Institute, Nanjing 210029, China.

*Corresponding authors: Ziwei Li (lzw@hhu.edu.cn); Junliang Jin (jljin@hhu.edu.cn)

## Abstract

Applying process-based Earth science models to new sites and questions often requires expert calibration decisions. Reliable delegation to language-model agents must preserve scientific competence and reproducibility. Here we develop a general calibration harness (GCH) for end-to-end delegation from a calibration request to assessed results. Constrained by a knowledge infrastructure (KI) and a validation convention, agents define parameter domains, observation mappings and objectives, and generate model-specific code. GCH checks the formulations against the same declarations and freezes them before numerical search and assessment, which require no further agent judgement. Agent-authored workflows executed across six model families, with assessment identifying unmet criteria. In snow-model tests, five of five knowledge-guided specifications executed on the first attempt, compared with none of five given only the model name, achieving validation Nash–Sutcliffe efficiency (NSE) of 0.81–0.94 against 0.57 for defaults. Under a common objective, independent GR4J formulations yielded scores within 0.001 NSE; re-evaluating eleven saved parameter vectors reproduced their metrics. In ten paired VIC tests with matched budgets, fixed search was non-inferior to continuing agent proposals (margin, −0.02 NSE; median difference, −0.003). This separation enables end-to-end calibration delegation with inspectable scientific choices and reproducible numerical evaluation.

## 1 Introduction

Process-based models of crops, catchments, snowpacks, land surfaces and aquifers are the instruments through which Earth and environmental scientists test process understanding and make predictions [1,2]. A model is built once and applied many times, and each application has its own scientific aim, spatial scale and observational situation [3]. A catchment model used to estimate total runoff, and the same model used to analyse runoff, snow storage and evaporation together, are one code applied under different constraints and judged against different targets. Bringing a model into a specific application therefore means matching its configuration to the available data, the question asked and the standard of evidence expected, and it usually means organizing a calibration and an evaluation for that application.

Numerical optimization executes an objective that has already been defined, and the definition rests on prior judgements. Which parameters may vary, and within what ranges, depends on the model configuration and on how far the observations can constrain them, because many parameter sets reproduce an observed series equally well [4,5]. Which simulated quantity corresponds to which observation, in what units and at what temporal and spatial resolution, decides whether a comparison is physically meaningful, and comparing several fluxes and states at once exposes compensating errors that a single series conceals [6,7]. Which metric expresses the question changes the parameter set that is selected [8,9], and the criteria by which a result is judged adequate are conventions of a field, codified for hydrological variables [10] and still debated for crop models [11]. Together these choices determine what the model is tested against. They are documented unevenly, in code comments, manuals, methods sections and the working notes of experienced users, and reassembling them for each new application is a large part of the effort of calibration.

Large language models can read and organize such material, generate the interface code that connects a model to data and optimizers, and act on simulation feedback. Each ability has been examined in Earth-science modelling. Language models have proposed parameter updates from simulation diagnostics inside an iterative calibration, where their proposals were compared with established optimizers [12]; they have orchestrated flood-modelling workflows from data retrieval to reporting under human oversight [13]; they have been trained as policies that learn parameter proposals from simulator feedback [14]; and they have reasoned within codified forecasting skills while numerical tools performed calibration and simulation [15]. Executable environments have made such work checkable in software repair and scientific coding [16,17], and knowledge infrastructures that package how a model is operated have allowed agents to run simulation workflows across models [18]. The parameter-update role of a language model has thus been tested. The role examined here comes before it: the formulation of the calibration experiment itself.

Whether existing model knowledge can support that role is an open question. A formulation written by a language model must satisfy two kinds of constraint at once. Interface constraints require that the named observation exists in the model output, that each parameter is applied in the unit and location the model expects, and that the metric is computed on comparable quantities. Scientific constraints require that the parameters, ranges, objectives and acceptance criteria fit the application at hand. A program can violate both and still run to completion, because a mismatched variable name or a parameter in the wrong unit leaves the numerics internally consistent while the experiment is wrong, and assessments of foundation models in hydrometeorology have documented such method-selection errors alongside successful programming [19]. The question is therefore whether the knowledge that defines a model's interface and its evaluation conventions can be supplied in a form that guides the generation of an experiment and that the same software then uses to check, execute and assess it, so that generation, execution and evaluation share one definition.

Here we test this with a general calibration harness (GCH) in which knowledge-guided language model agents formulate calibration experiments and deterministic software executes them. The agent reads a model's knowledge infrastructure [18], extended for calibration with a model graph that declares what each output can be compared with and a validation convention that records declared, literature-informed assessment criteria, and writes a calibration contract and the model-specific runner once, before any search. GCH compiles objectives from the same declarations, checks the generated code against the model's reference run, performs numerical search with conventional optimizers and assesses every observed variable against the declared criteria. We ask whether agents formulate executable experiments for six model families with different interfaces and observation structures; whether the supplied knowledge is what makes the formulations usable, using comparisons across information conditions for a snow model; and what fixed formulations deliver numerically, using repeated formulations, a paired comparison between fixed search and continuing agent proposals, and replays of saved evaluations. Knowledge-guided formulations executed across the six families and the assessment identified which observed targets remained unresolved; model-specific interface knowledge decided first-pass execution; and fixed formulations gave closely similar results under a shared objective and met the non-inferiority criterion without agent calls during search. Each calibration leaves an explicit, reusable experiment definition in which the scientific choices of a model application can be inspected and compared.

# 2 Methods

## 2.1 Knowledge infrastructure and the general calibration harness

A knowledge infrastructure is an inspectable package that encapsulates how a model is operated [18]. Each package used here contains the nine components listed in Table B1: an operating manual; staged protocols for forcing, configuration, execution, parsing, validation and calibration; validated operators, including the runner that a specification invokes; diagnostic triplets of symptom, diagnosis and remedy; the model graph; the validation convention; an input–output specification; a literature index; and a manifest recording the measured skill of the reference run. The agent reads the procedural components. The model graph and the validation convention are declarative inputs read by both the agent and the harness; they are written when the package is built and remain fixed through calibration.

The model graph is synthesised from five sources read separately: the source code, the official documentation, the existing operating package, the sensitivity and calibration literature and the recorded failure triplets. Each reader is confined to one source and emits structured facts. A synthesiser merges the five streams into an eight-block template (identity, boundary, inputs, processes, states, outputs, influence, safety) under a fixed precedence in which the code decides what the model emits, the documentation decides what it declares and every influence edge requires a citation. One verifier per block returns a pass or revise verdict and the synthesiser runs at most once more. The resulting file declares, for each output, the observation types it can be compared with, the metric families that apply and the determining metric. The validation convention records, per output variable and observation type, the metric, its direction, the numeric bands and the citation. Papers are gathered from a local cache and a capped open-access search;

an extraction agent writes the entries, or a null band with low confidence where no paper gives a number; a script checks that every non-null band cites a fetched paper and returns violations for at most two retries; every band is then read against its source by three readers, one standard per quantity is reconciled across models and the file is frozen. Because every metric carries a direction and every band is a number, the harness applies the bands without domain knowledge (Fig. A1).

The harness loads the contract and the graph, compiles objectives from the metric families that the graph declares for each target, and rejects a specification that names a variable the graph does not list as an output or a metric family it does not declare. Before search it certifies the runner by injecting the validated reference parameters of the package and requiring the headline metric to reproduce the reference run within a tolerance of 0.02, together with the recorded run signature where both sides report one. On every evaluation it compares the requested parameter vector with the runner's applied-parameter report, and it treats any mismatch, failed execution or non-finite metric as an infeasible loss. At commissioning it perturbs each parameter and requires the metric to respond. Parameters are injected either by a generic applicator into stable input files or, where the model regenerates its inputs at run time, by the runner itself, which reads the values back from the files the model consumes. Scalar search uses SPOTPY [30] with DDS [21], SCE-UA [27] or DREAM [22]; multiobjective search uses pymoo [31] with NSGA-II [23]; a scalar optimizer requested for a multiobjective task is replaced. Holdout assessment follows the split declared in the contract. The reported experiments used the archived version of the harness; the current release adds measured-cost budgeting and optimizer-specific convergence diagnostics (Appendix C).

## 2.2 Calibration contracts and runners

The contract is a machine-readable file that declares the case identity, the observation contract, the targets, the parameters (name, description, units, default, range, hard bounds, transformation, spatial scope and tying, and the file address or runner key through which each is applied), the constraints, the objectives, the search strategy, the budget and the holdout. The runner is a program that receives a candidate vector, writes it into private copies of the model's native inputs, executes the model binaries, scores the declared targets and emits the metrics together with its applied-parameter report, case identity and run-health counts; on any failure it emits nothing and exits with an error, and the harness records an infeasible loss. Agents authored both files from the knowledge package and the contract schema. Preparation comprised generation, review by an independent model or commissioning runs, and revision where commissioning required it; each formal search then executed the fixed formulation. The continuing arm of the paired VIC experiment is the one search in which an agent proposed parameters, and it exists to test that difference. For FSM2 and the GR4J range tests a fixed runner was supplied and the agent wrote only the contract.

## 2.3 Assessment and component acceptance

Each observed variable is assessed by metric family. The temporal loss is max(0, 1 − NSE) [20]; the magnitude loss is absolute percentage bias divided by 100, except for the declared-budget WOFOST application, which uses root-mean-square error normalized by the mean observation;

the HBV snow-timing loss is the absolute mean signed annual peak-date bias divided by 30 days. Discharge criteria follow NSE above 0.50 and absolute bias below 15% [10], and crop assessment adopts a 30% normalized-error ceiling [32]; snow and evapotranspiration thresholds are literature-informed operational choices recorded in each convention, and comparisons are inclusive at the recorded ceilings. A component is accepted when its calibration and validation losses are finite, its validation loss meets the component-specific ceiling or the applicable fallback, its Pearson correlation is at least 0.50 in both periods for temporal components, and its validation loss does not exceed the uncalibrated baseline by more than $10^{-6}$. Where the convention stores no ceiling for a component, the component is decided by a fallback magnitude ceiling of 30% absolute bias where applicable and otherwise by the correlation and baseline tests; the ceilings are those stored in each convention when the campaign ran, and Table B9 lists the rule that decided every component. A workflow is accepted only when every component is accepted. In the six-family application, final-front members were ranked by their largest normalized calibration loss, excluding objectives with negligible spread, and assessed in rank order until one passed every component; where none did, HBV, VIC and CRHM selected among discharge-acceptable members by minimizing the largest normalized validation loss of the remaining components, and WOFOST and SUMMA retained the first-ranked member. Validation data of the six-family application therefore informed solution selection, whereas FSM2 and the paired VIC experiment selected parameters on calibration data alone.

### 2.4 Six-family applications

WOFOST simulated maize at Kellogg Biological Station, where grain mass and destructive aboveground biomass are observed, for ten years within 1991–2017 alternated into five calibration and five validation years, with seven parameters. HBV simulated the 239 km$^2$ Moyie catchment with discharge from HYDAT station 08NH120 and snow water equivalent from the CanSWE pillow BCE-2C10P, with ten parameters calibrated over 2006–2010 and validated over 2011–2015; SUMMA used the same observations and periods with one grouped response unit, three hydrologic response units and nineteen parameters. VIC simulated the roughly 10,190 km$^2$ Xixian catchment with CMFD forcing, discharge from gauge 50225601 and monthly GLEAM 4.2a evapotranspiration, calibrated over 2007–2008 and validated over 2009–2010, with thirteen active parameters from a pool of fifteen. MODFLOW 6 used the Sagehen example distributed with the model, with four parameter multipliers and discharge from USGS gauge 10343500; because the site has no head observations, the targets were streamflow and baseflow duration curves at nineteen exceedance probabilities, compared with the two halves of a 26,624-day observed record split on 13 March 1990, with baseflow derived by the Lyne–Hollick filter. CRHM simulated the 272 km$^2$ Blue River catchment with discharge from HYDAT 08LB038 and snow water equivalent from CanSWE BCE-1E02P, calibrated over 2006–2010 and validated over 2011–2015, with 26 parameters including precipitation correction.

In the single-objective phase, each family was calibrated once for one observed variable (grain mass for WOFOST, discharge for HBV, SUMMA, VIC and CRHM, and streamflow for MODFLOW 6). The agent read the knowledge package and authored the contract and runner; each specification was reviewed by an independent model before execution and was then executed unchanged at a budget of 80 evaluations. Five specifications selected DDS and CRHM selected SCE-UA. Poor-baseline retests used DDS at 120 evaluations for HBV and SUMMA and three seeds at 600 evaluations for CRHM at Belly River. Authoring cost was measured from the command-line usage record for the six authorings, and the WOFOST, HBV and MODFLOW 6 tasks were re-authored by three further configurations from two vendors (Appendix D). The working directory of these runs was lost to a server reboot; their specifications are retained and their numerical results survive as the console record of the coordinating session (Appendix D), so Tables B3–B5 report them descriptively and they enter no figure.

In the multivariable phase, a coordinating agent and model-specific agents prepared the six formulations under human direction of the scope of the campaign and the choice of cases. Commissioning runs of 40–200 evaluations tested each formulation before the declared-budget campaign. All six searches used NSGA-II with seed 0 at the nominal budgets declared in the contracts (600 for WOFOST, 500 for HBV, 240 for SUMMA, 150 for VIC, 200 for MODFLOW 6 and 500 for CRHM); population termination produced 520 optimizer evaluations for HBV and CRHM. WOFOST activated three, five and seven parameters in successive stages and VIC increased its active set from three to thirteen parameters across six stages. The saved results contain 2,536 distinct calibration vectors, 385 validation records and 167 final-front members.

## 2.5 Knowledge experiments with FSM2 and GR4J

The preregistered FSM2 experiment used the Mt Rose site (SNOTEL 652) with hourly forcing in the native driving format of the model and daily observed snow water equivalent from the co-located pillow. Three information conditions supplied the structured knowledge infrastructure (twenty files, 226,250 bytes), the raw documentation of the model (nine files, 92,822 bytes) or the model name together with the common task schema; each condition had five independent generation sessions, each of which produced a specification. The common task fixed the calibration window (water years 2016–2018, 1,096 days) and the validation window (water years 2019–2020, 731 days), required the objective to be NSE, KGE or correlation, and fixed the optimizer as DDS with seed 0 at a budget of the agent's choosing between 6 and 2,000 proposals; generated budgets were 200–500. The compiled defaults gave validation NSE 0.57 and KGE 0.43. The primary outcomes were execution without revision and validation NSE above the defaults; the support criterion required at least four of five K and at most one of five M specifications to execute. The repair protocol first corrected the target name from SWE to the exposed variable snw, then converted the hourly timescales tmlt, tcld and trho to seconds by multiplication by 3,600; specifications containing the unsupported parameter bstb received no repair search. The experiment comprised fifteen generations and 21 searches, with 6,350 proposals, 6,348 scored proposals and 5,653 completed model executions. An exploratory precursor with three generations per condition informed the confirmatory design and was analysed separately (Appendix E).

For GR4J, nine specifications (three per condition) were assessed without calibration. Each proposed parameter range was transformed to the canonical coordinates of the model and compared by intersection over union with the empirical 10th–90th percentile box of the developers [25]: X1 in [100, 1,200] mm, X2 in [−5, 3] mm $d^{-1}$, X3 in [20, 300] mm and X4 in [1.10, 2.90] days; the overlap was averaged across the four parameters for each specification.

## 2.6 Repeated formulations and paired search

Five fresh knowledge-guided GR4J formulations were executed with optimizer seeds 0, 1 and 2 on a common runner over the full 1990–1999 window (3,595 valid days). Four selected SCE-UA and one selected DDS, with nominal budgets of 1,500–3,000 evaluations; SCE-UA terminated after 1,302–2,135 evaluations. Two formulations selected KGE as their objective, and each of their seeded searches was also executed with NSE substituted, giving six matched interventions and 21 searches. The consistency analysis used the fifteen NSE endpoints, nine from the original NSE formulations and six from the interventions.

The preregistered VIC experiment at Tangnaihai fixed a six-parameter formulation with loss 1 − NSE, calibration over 2007–2011 and validation over 2012–2016. Each of ten paired blocks shared five starting points, the runner default and four Latin-hypercube points. The fixed arm generated 145 further proposals with DDS and no language-model calls; the continuing arm obtained 145 accepted proposals from an agent that received the running history of parameters and diagnostics and returned the next vector, with 1–18 rejected proposals per block. Both arms executed exactly 150 calibration evaluations and one validation evaluation of the calibration-selected vector. The estimand was the median paired difference in validation NSE, fixed minus continuing, with a non-inferiority margin of −0.02. Ten continuing-arm attempts interrupted by transport errors, rate limits or a host reboot were excluded under the preregistered exclusion rules; no fixed-arm attempt failed. The proposer is a compact diagnostic proposer built for this comparison; it implements the parameter-update role studied in ref. 12 without reproducing that study's complete workflow.

## 2.7 Reference comparisons

The generated GR4J workflow was evaluated at basin L0123001 (360 $km^2$) with 1989 as warm-up and 1990–1999 as the scoring window. DDS with seed 0 used 1,000 evaluations; the airGR Calibration_Michel reference [28] used 234. The generated parameters were 253.65, 0.97, 87.55 and 2.21 against reference values of 257.24, 1.01, 88.23 and 2.21. The generated SAC-SMA workflow for the Blanco River (USGS 08171300) used Maurer forcing, Hamon potential evapotranspiration, eleven free and five fixed parameters, and SCE-UA with seed 0, which used 2,702 of a declared 3,000 evaluations; calibration from October 1999 to September 2008 gave NSE 0.72, KGE 0.66 and bias +26.2%, and validation from October 1989 to September 1999 gave NSE 0.70, KGE 0.69 and bias +24.8%. The CAMELS benchmark [29] couples Snow-17, SAC-SMA and a unit hydrograph with a different evapotranspiration scheme. Fitted parameters were compared with ten Daymet-forced reference parameter files for the basin; ten parameter identities matched and the eleventh, uzk, is constant across the files (Fig. A5).

### 2.8 Numerical comparisons and replays

SAC-SMA at Ibex (HYDAT 09AC007) used five equal-area hydrologic response units, degree-day snow and Hamon potential evapotranspiration, simulated over 2005–2015 with calibration over 2006–2010 and validation over 2011–2015 and thirteen free parameters, in a hand-written, literature-derived setup. DDS, SCE-UA and uniform random search were run at nominal budgets of 300, 1,000 and 3,000 with seeds 1, 2 and 3, giving 27 searches and 35,566 finite evaluations; the calibration-best vector was selected before validation scoring. SCE-UA realized 388–392, 907–919 and 1,935–1,963 evaluations at the three budgets (Fig. A4). HYMOD at Leaf River, weighted-scalar against multiobjective searches for HYMOD and WOFOST, parameter-range and selection-rule diagnostics for the six families and a sampling adapter demonstration are described in Appendix G (Fig. A6). Eleven replay packages each evaluated one saved parameter vector with the archived environment and compared the numerical outputs with the saved record (Table B12). Tables B1–B24 list the individual numerical results.

### 2.9 Language-model configurations

Initial single-objective authoring used claude-fable-5; the six-family multivariable campaign, the FSM2 and GR4J knowledge experiments, the repeated GR4J formulations and the reference comparisons used claude-opus-4-8; the continuing proposer in the paired VIC experiment used claude-sonnet-5; cross-vendor authoring used claude-sonnet-5, claude-opus-4-8 and GPT-5.5. The agents used pretrained language models without task-specific fine-tuning. Agent sessions ran through the Claude Code command-line interface (version 2.1.258) in sandboxes with read-only access to the knowledge package and the contract schema and with no state carried between sessions; the interface exposes no sampling parameters, so independent sessions serve as replicates. Every prompt, transcript, contract, runner and evaluation is retained, except the numerical outputs of the single-objective phase noted above. OpenAI Codex and Anthropic Claude also assisted with evidence review, manuscript development and figure preparation; the authors are responsible for the study and the final text.

### 2.10 Statistical analyses

The FSM2 comparison of first-pass execution between the K and M conditions used a two-sided Fisher exact test on the five-by-five outcome ($P = 7.9 \times 10^{-3}$); the preregistered decision rule was the support criterion stated above. The paired VIC analysis estimated the median paired validation-NSE difference across the ten blocks, with a 95% interval from 10,000 complete-block bootstrap resamples (seed 0) and the preregistered non-inferiority margin of −0.02 applied to the lower bound. Replication units are generation sessions for the knowledge experiments, paired blocks for the VIC comparison and optimizer seeds for the GR4J and SAC-SMA comparisons; all other comparisons are descriptive, and the number of searches, evaluations or components is stated with each result.

# 3 Results

## 3.1 Knowledge-guided formulation across model families

Figure 1 shows the division of work. A calibration task names a model, the observations and the targets. The agent reads the knowledge infrastructure of the model, which packages staged operating protocols, validated tools and recorded failure modes [18], together with two declarations added for calibration. The model graph lists every input, parameter and output with its units and, for each output, the observation types it can be compared with and the applicable metric families. The validation convention records, for each variable and observation type, the metric, its direction and the acceptance band used in the model's own literature, with its source. Both files are built once per model from the source code, documentation, operating package and calibration literature (Fig. A1; Methods). The agent's product is a calibration contract, which declares the targets, the parameters with their transformations, bounds and constraints, the objectives, the search strategy, the budget and the holdout, and a runner, which writes a candidate parameter vector into the model's native inputs, executes the model and returns the metrics with a report of the parameters it applied. GCH compiles objectives only from target–metric combinations that the graph declares, reproduces the reference run of the package through the generated runner, compares requested and applied parameters at every evaluation and confirms at commissioning that each parameter changes the output; it then searches with conventional optimizers and assesses each observed variable against the declared criteria. Generation, checking and revision form a preparation stage; the formulation is fixed before formal search, which runs without language-model proposals (Table B1).

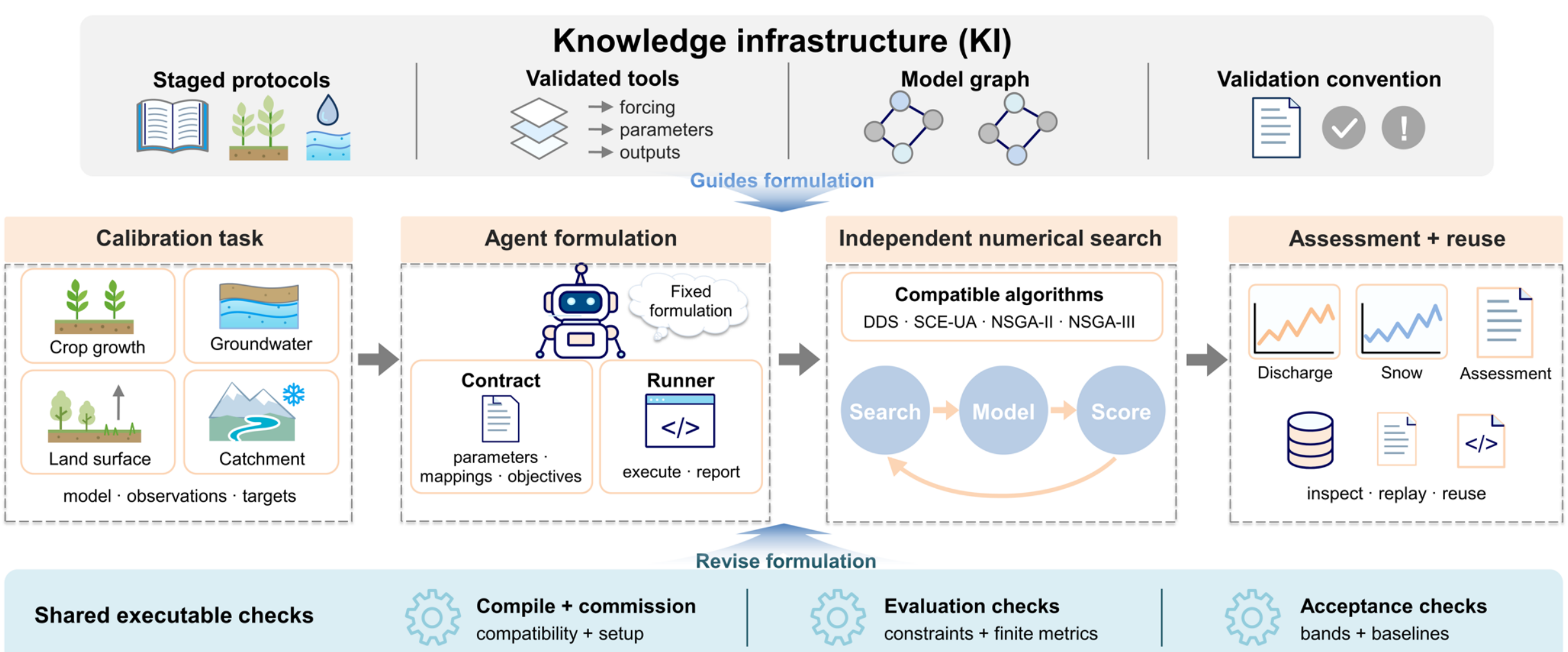

**Fig. 1 Knowledge-guided formulation of calibration experiments and their numerical execution.** A calibration task (left) names a model, its observations and its targets. The knowledge infrastructure of the model (top) supplies staged protocols, validated tools, a model graph and a validation convention. Reading them, the agent (centre left) writes a fixed formulation: a contract declaring parameters, observation mappings and objectives, and a runner that executes the model and reports metrics. The general calibration harness (centre right) searches the declared parameter domain with compatible optimizers through repeated search, model and score cycles, and assessment and reuse (right) judge each observed variable against the declared criteria and retain the contract, the code and the records for inspection and replay. Shared executable checks (bottom) compile and commission the formulation, check constraints and finite metrics at every evaluation, and apply acceptance bands and baselines; their feedback returns to the agent only during preparation, to revise the formulation.

Six model families from the knowledge library received the same kind of request, to calibrate the model against the available observations, under different interfaces and observation structures (Table 1; Table B2). WOFOST simulated maize at Kellogg Biological Station, where grain mass and destructive aboveground biomass are measured; HBV and SUMMA simulated the snow-dominated Moyie catchment, with discharge and snow water equivalent from a co-located pillow; VIC simulated the Xixian catchment with discharge and monthly satellite-derived evapotranspiration; MODFLOW 6 used the Sagehen example distributed with the model, where a gauge but no head observations exist, so that streamflow and filtered baseflow duration curves were the targets; and CRHM simulated the Blue River catchment with discharge and snow water equivalent. Each task required its own parameter set and injection route, observation mapping, objective and budget.

**Table 1 Experiments, agent tasks, replication units and assessment.** Paired periods denote calibration and validation. Appendices D–G give the full protocols and retained records, and Table B2 lists the evaluated outcomes and supporting material of every experiment.

| Experiment | Question | Model and case | Agent task and preparation | Replication unit | Assessment |
|---|---|---|---|---|---|
| Single-objective formulation | Do knowledge-guided formulations execute for different model interfaces? | WOFOST, Kellogg Biological Station; HBV and SUMMA, Moyie; VIC, Xixian; MODFLOW 6, Sagehen; CRHM, Blue River | Specification and runner written from the knowledge package; review by an independent model; executed unchanged at 80 evaluations | One formulation per family (6) | Execution; holdout criterion; retests at sites with poor defaults |
| Cross-configuration re-authoring | Does execution depend on the authoring model? | WOFOST, HBV and MODFLOW 6 single-objective tasks | Re-authored by three further language-model configurations from two vendors | One authoring per task and configuration (9) | Runnable specification; optimizer choice; cost and time |
| Multivariable calibration | What do formulations deliver for two observed quantities? | The six cases above, two observed variables each (WOFOST, ten years within 1991–2017 alternated; HBV, SUMMA and CRHM, 2006–2010 and 2011–2015; VIC, 2007–2008 and 2009–2010; MODFLOW 6, record split on 13 March 1990) | Coordinating and model-specific agents; commissioning at 40–200 evaluations; NSGA-II at declared budgets of 150–600 evaluations | One search per family (6); 23 variable–metric components | Completion; joint acceptance rule with validation-guided member selection |
| Information conditions | Which supplied knowledge decides executability and skill? | FSM2, Mt Rose (water years 2016–2018 and 2019–2020); GR4J parameter ranges, static | Specification only, runner supplied; structured knowledge, raw documentation or model name; preregistered; prescribed target and unit repairs | Independent generation sessions (5 × 3 for FSM2; 3 × 3 for GR4J) | First-pass execution; validation NSE of calibration-selected vectors; overlap of proposed ranges with the empirical domain |

| Experiment | Question | Model and case | Agent task and preparation | Replication unit | Assessment |
|---|---|---|---|---|---|
| Repeated formulation | How much do independent formulations differ under a shared objective? | GR4J, basin L0123001 (1990–1999) | Five formulations on a common runner; six KGE-to-NSE interventions | Formulation × optimizer seed (15 NSE endpoints; 21 searches) | Spread of fit; effect of the objective |
| Fixed search against continuing proposals | Does numerical search need an agent? | VIC, Tangnaihai (2007–2011 and 2012–2016) | Fixed six-parameter formulation; DDS alone against agent proposals from running diagnostics; shared starting vectors; preregistered | Paired block (10); 150 calibration evaluations per arm | Median paired validation NSE difference against a −0.02 non-inferiority margin |
| Numerical replay | Can saved evaluations be reproduced? | Eleven packages across the applications above and the reference cases | One saved parameter vector re-evaluated in the archived environment | Package (11) | Agreement with the saved metrics |
| Reference comparison | How do generated workflows compare with established calibrations? | GR4J, L0123001, against airGR; SAC-SMA, Blanco River, against CAMELS and ten reference parameter files | Generated workflows under supplied case information | One calibration each | Fit, bias and parameter-domain differences |

In a single-objective phase, one agent read each package and wrote a specification and runner for one observed variable; each specification was reviewed by an independent language model and then executed unchanged at 80 evaluations (Table B3). All six executed. WOFOST, VIC and MODFLOW 6 completed calibration and met their holdout criterion, and HBV, SUMMA and CRHM stopped before optimization because the harness found that the default parameters already met the declared criterion at those sites. Applied to basins with poor default fit, the same formulations raised holdout Nash–Sutcliffe efficiency (NSE) [20] from 0.59 to 0.73 for HBV, from 0.40 to 0.61 and 0.70 for two SUMMA seeds and from 0.44 to 0.71–0.73 for CRHM. Re-authoring the WOFOST, HBV and MODFLOW 6 tasks with three further language-model configurations from two vendors produced nine further runnable specifications (Table B4). Generating one specification and runner with the first configuration used 16,575–87,042 output tokens, cost US$3.57–14.21 and took 11–34 min; these figures cover generation alone and exclude the construction of the knowledge package, data preparation and the model runs.

The contracts record the decisions that the knowledge supported. The VIC contract seeded its parameter pool from the influence edges of the model graph, retained the parameters that the sensitivity literature of the model identifies as dominant, excluded adjustments to the meteorological forcing on the grounds that they correct inputs rather than calibrate the model, omitted a canopy parameter because the generated input files contain no field for it, and narrowed the soil-layer depth ranges so that every point of the search box satisfies the model's own feasibility constraint. Optimizer choices followed the structure of the request: HBV, VIC and MODFLOW 6 selected dynamically dimensioned search (DDS) [21] when the deliverable was a point estimate and the DREAM sampler [22] when it was parameter uncertainty, and in a blinded single-shot test the models of three vendors agreed on twelve of twelve optimizer choices across four task framings (Table B5). The checks also corrected choices: in the multivariable phase described below, the SUMMA and VIC specifications requested DDS for a two-objective task, and GCH replaced it with the non-dominated sorting genetic algorithm NSGA-II [23]. Appendix D reports the single-objective, re-authoring and multivariable records.

### 3.2 Model knowledge resolves interface errors

To isolate what the supplied knowledge contributes, agents received the same calibration task for the Flexible Snow Model 2 (FSM2) [24] at a mountain snow-pillow site under three information conditions, with the runner supplied so that only the specification varied: the structured knowledge infrastructure (K), the raw documentation of the model (R) or the name of the model with the task schema alone (M). Each condition had five independent generation sessions, and every search used DDS with seed 0 at the budget each specification declared (Methods). All five K specifications executed on the first attempt, compared with three of five R and none of five M specifications (Fig. 2a; two-sided Fisher exact test of K against M, $P = 7.9 \times 10^{-3}$). K calibrations reached validation NSE of 0.81–0.94 against 0.57 for the compiled defaults (Table B6).

The errors were concentrated in two interface facts. Every M specification named the observation target SWE, the physical variable, where the runner exposes it as snw; the specification was physically sensible but could not be executed. Every R and M specification expressed the albedo and compaction timescales in hours, where the interface requires seconds; these searches ran, but the domain they explored was 3,600 times too short, so the optimizer settled on melting-snow albedo timescales of minutes, where the physical scale is days (Fig. 2b). Two M specifications additionally requested a parameter (bstb) that the model does not have.

Repairing only these two decisions, with every other choice unchanged, made all five R and the three eligible M specifications runnable, and the unit correction raised validation NSE in all eight eligible pairs, to 0.96–0.98 for R and 0.93–0.99 for M (Fig. 2c). The paired repairs thus identified target mappings and parameter units as the sources of formulation error in this experiment; the objectives, parameters and budgets chosen without the knowledge were serviceable once those two facts were supplied. The repaired R and M calibrations exceeded the first-pass K range because their specifications had declared different objectives and budgets. The K and R packages differed in content and volume as well as in organization, so the experiment attributes first-pass execution to the model-specific interface knowledge they supplied (Appendix E).

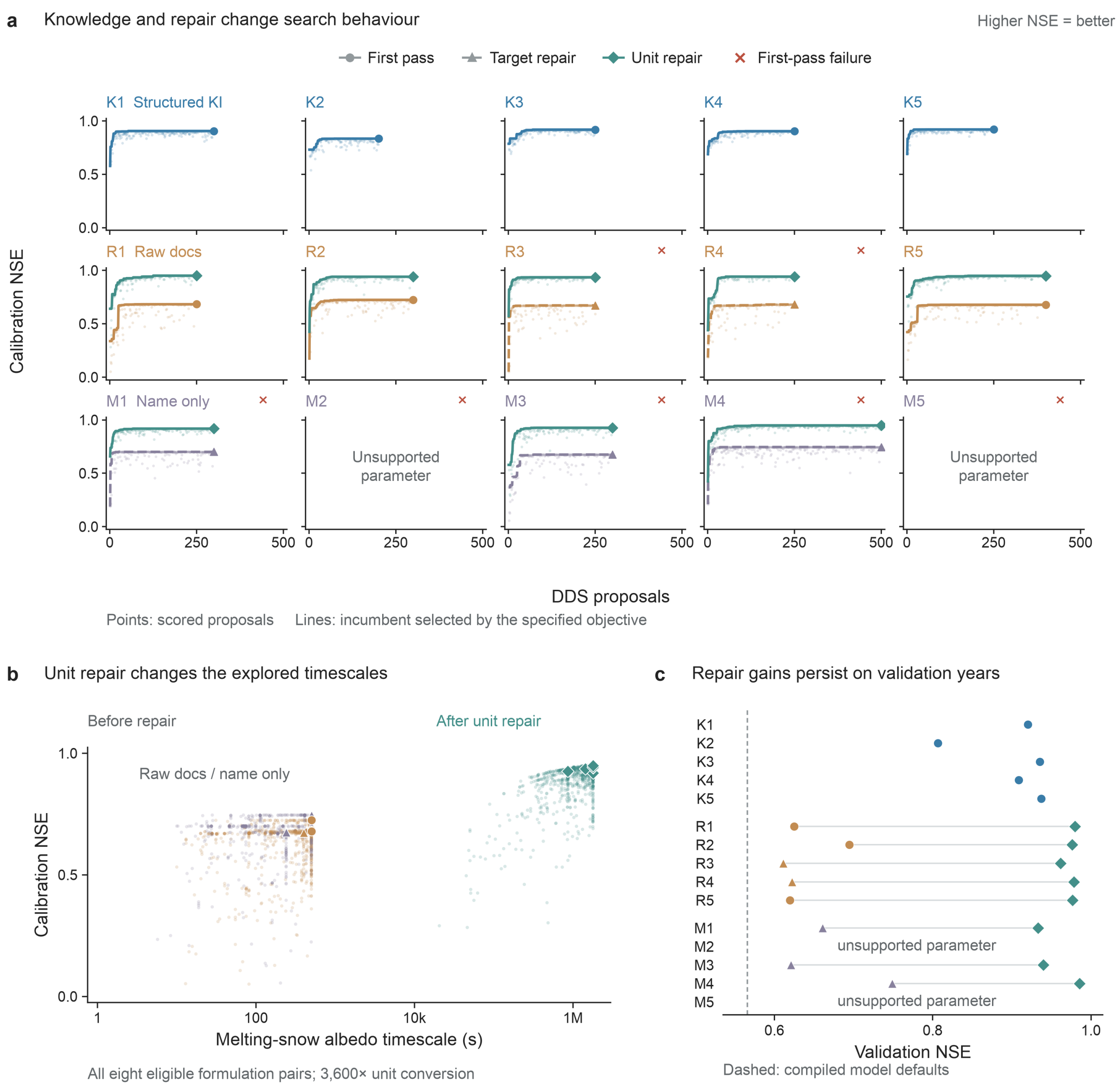

**Fig. 2 Supplied knowledge, interface errors and prescribed repairs in FSM2. a**, Calibration NSE against DDS proposals for fifteen independent specifications written with the structured knowledge infrastructure (K1–K5), raw documentation (R1–R5) or the model name only (M1–M5). Points show scored proposals; lines track the incumbent selected by each specification's own objective; circles, triangles and diamonds mark the end of first-pass, target-repaired and unit-repaired searches; red crosses mark first-pass failures; panels labelled unsupported parameter received no repair search. **b**, Calibration NSE against the melting-snow albedo timescale for all eight eligible formulation pairs before (orange and purple) and after (green) the 3,600-fold unit conversion; large markers show the selected vectors. **c**, Validation NSE of the 21 selected vectors, with repair stages linked within each specification; the dashed line marks the compiled model defaults (0.57). Calibration used water years 2016–2018 and validation water years 2019–2020.

A static test with the GR4J rainfall–runoff model [25] gave the same picture at the level of parameter ranges. Nine fresh specifications, three per condition, were assessed without calibration against the empirical parameter domain of the model's developers. Ranges proposed with the knowledge infrastructure overlapped that domain by 0.81–1.00 (intersection over union averaged over the four parameters), compared with 0.41–0.53 from raw documentation and 0.39–0.60 from the name alone; all nine ranges contained the reference optimum, and every name-only specification again selected an observation target that the runner does not expose (Fig. A2).

### 3.3 Multivariable assessment identifies unresolved targets

The multivariable phase asked each of the six models to match two observed quantities at once: grain mass and biomass for WOFOST, discharge and snow water equivalent for HBV, SUMMA and CRHM, discharge and evapotranspiration for VIC, and streamflow and baseflow duration curves for MODFLOW 6. The formulations were commissioned at 40–200 evaluations (Table B7) and then ran with NSGA-II at the budgets the agents had declared, 150–600 evaluations. All six searches completed (Fig. 3). Final Pareto members were selected with validation losses (Methods), so these results describe execution, the trade-offs among candidates and the declared assessment; held-out predictive skill is tested in the FSM2 and paired VIC experiments. Across the 23 variable–metric components, 16 met the joint acceptance rule: 0 of 2 for WOFOST, 4 of 5 for HBV, 3 of 4 for SUMMA, 2 of 4 for VIC, 4 of 4 for MODFLOW 6 and 3 of 4 for CRHM (Fig. 4). Counted by workflow, one of the six, MODFLOW 6, satisfied every component.

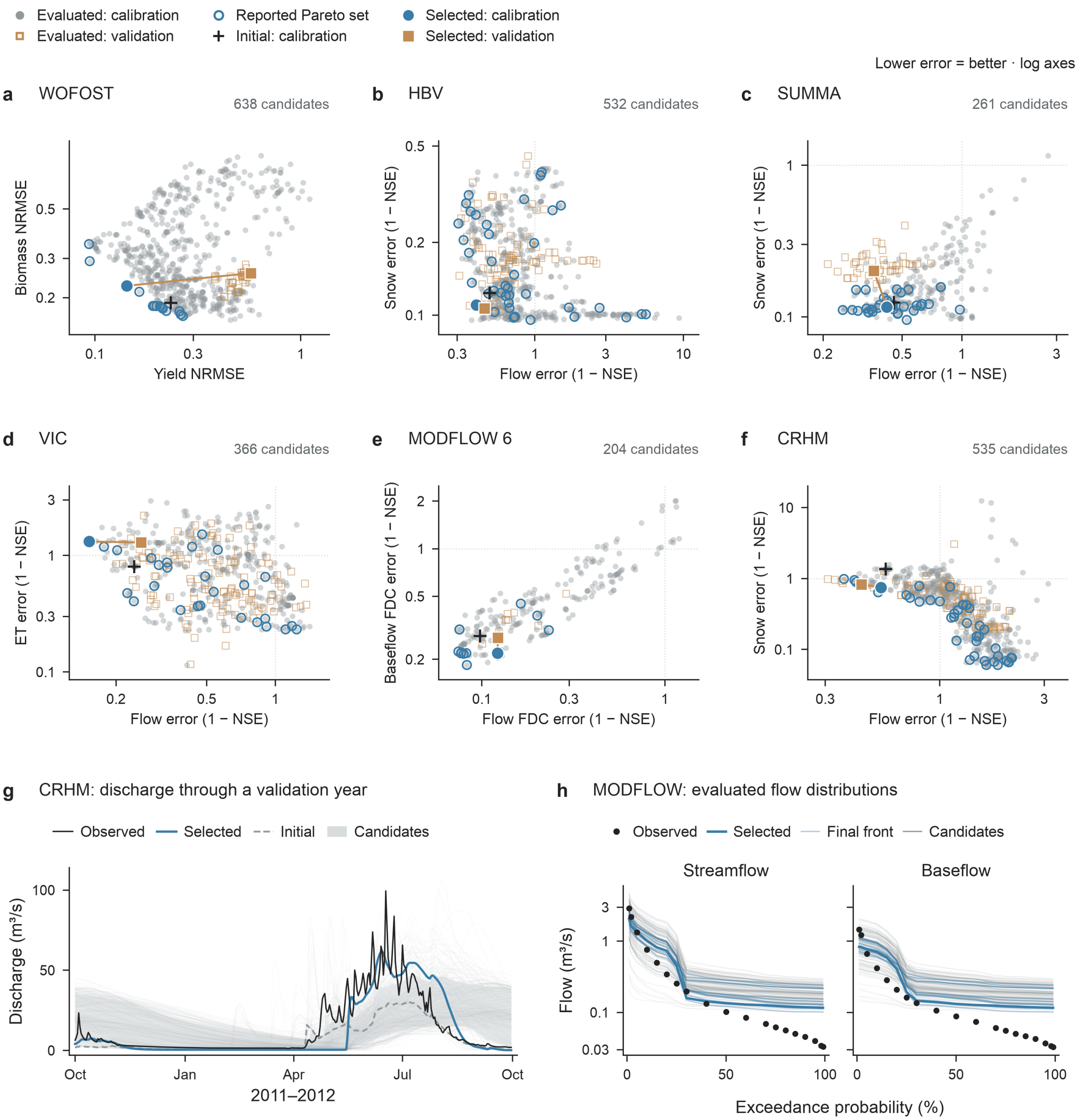

**Fig. 3 Multivariable calibration across six model families. a–f**, Objective space for WOFOST, HBV, SUMMA, VIC, MODFLOW 6 and CRHM. Grey circles, evaluated calibration vectors ($n = 2{,}536$ in total); open squares, validation assessments ($n = 385$); blue rings, members of the final Pareto front ($n = 167$); filled circle and filled square, calibration and validation scores of the selected member, joined by a line; cross, the uncalibrated starting point. Axes show normalized root-mean-square error for WOFOST and $1 - \mathrm{NSE}$ otherwise, on logarithmic scales; lower values indicate better fit; the number of candidates is given in each panel. **g**, Observed CRHM discharge (black) through the 2011–2012 validation year with the selected (blue) and initial (dashed grey) simulations; grey shading spans the 10th to 90th percentiles of the evaluated candidate responses. **h**, MODFLOW 6 streamflow and baseflow duration curves at nineteen exceedance probabilities: observed (dots), selected (blue), final-front members (light blue) and candidates (grey). Fig. 4 shows the component-level assessment.

The component results show how the two observed quantities interacted. MODFLOW 6 improved the distributional fit of both duration curves, reducing absolute bias from 34% to 4% for streamflow and from 56% to 12% for baseflow; both curves derive from one gauge record through a baseflow filter, so the two fits summarize a single observation. VIC retained its discharge skill, with validation NSE 0.74 and absolute bias reduced from 11% to 4%, while its evapotranspiration validation NSE fell from 0.05 to −0.29 and its evapotranspiration bias rose from 25% to 31%. WOFOST reduced calibration-period grain-mass error to 14%, but its validation error rose to 57%, above the 49% of the defaults, and its biomass error of 26% remained below the 30% ceiling yet above the default 21%. In the two snow–discharge cases, HBV raised validation discharge NSE from 0.43 to 0.54 and reduced discharge bias from 29% to 13% while its snow components stayed at the level of the defaults (snow NSE 0.89 against 0.90), and CRHM reduced discharge bias from 48% to 11% and raised snow NSE from −0.27 to 0.18 while snow bias remained at 58%.

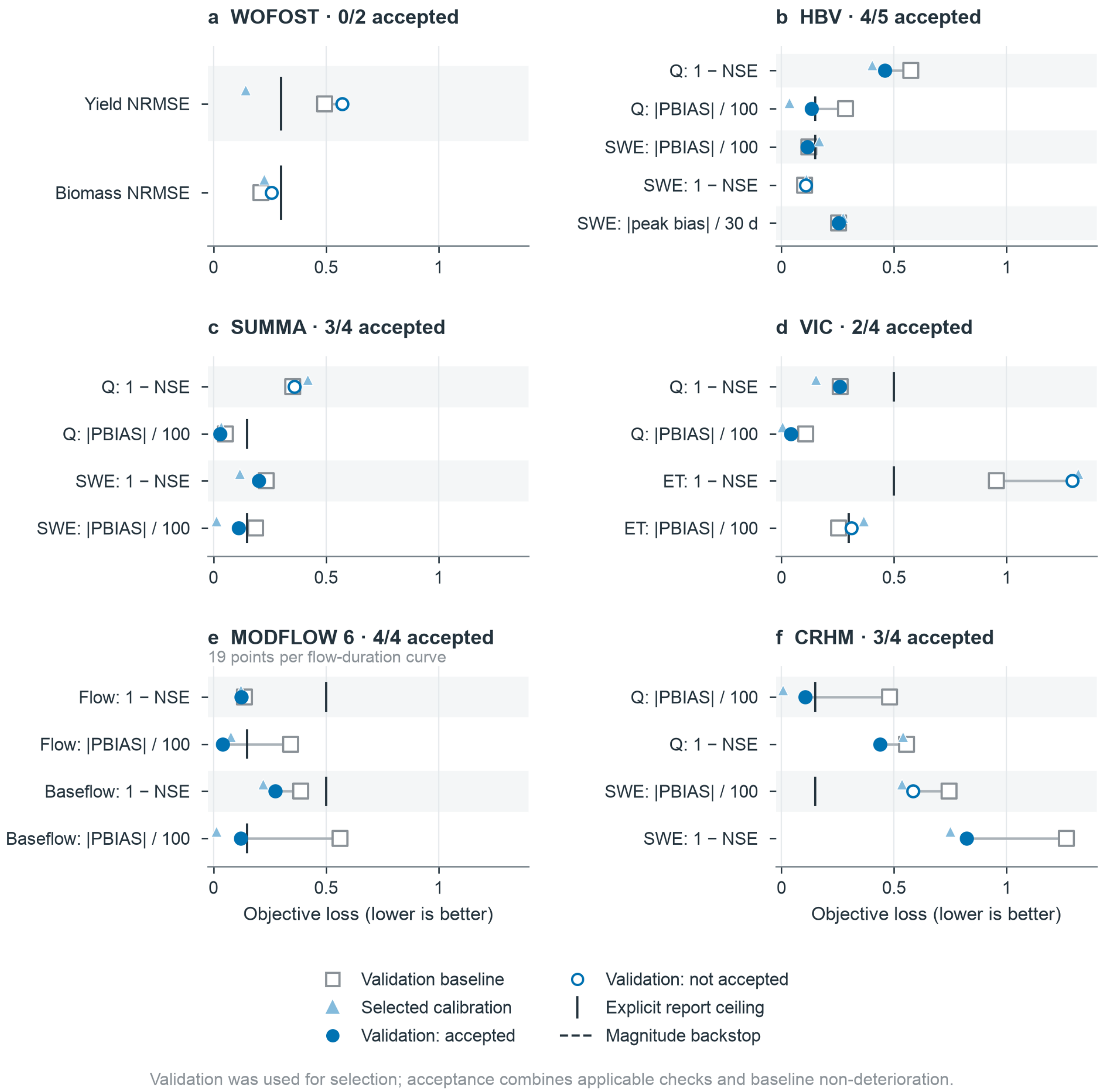

**Fig. 4 Component-level assessment of the six multivariable calibrations. a–f**, For each model, every variable–metric component is shown as a loss, with lower values better: open square, validation loss of the uncalibrated baseline; light-blue triangle, calibration-period loss of the selected member; filled blue circle, validation loss of an accepted component; open blue circle, validation loss of a component that was not accepted. Vertical bars mark the acceptance ceilings stored in the validation convention; a dashed bar marks the fallback magnitude ceiling applied where no ceiling is stored. Acceptance combines the applicable ceiling, the correlation and finite-loss checks and no deterioration from the baseline, so a component can fail at a loss below its ceiling (a deterioration from the baseline) or pass without a stored ceiling (an improvement over the baseline). Q, discharge; SWE, snow water equivalent; ET, evapotranspiration; FDC, flow-duration curve; PBIAS, percentage bias. Accepted counts appear in each panel title.

The joint rule combines the ceiling stored in the validation convention where one exists, a fallback magnitude ceiling where none is stored, correlation checks for temporal components and no deterioration from the uncalibrated baseline on the validation period (Methods; Tables B8 and B9). The WOFOST grain-mass component failed its ceiling and the baseline test, and its biomass component met the ceiling but failed the baseline test. The HBV snow temporal component and the SUMMA discharge component were rejected on the baseline test alone, at NSE 0.89 and 0.64, whereas the CRHM snow temporal component was accepted on the baseline test alone, at NSE 0.18, because the retained convention stores no ceiling for that variable. Under the tested formulations, the assessment located unresolved discrepancies in crop production for WOFOST and in evapotranspiration for VIC, and it names the variable and the criterion in each case. Widening the parameter ranges threefold or changing the selection rule on the same fronts altered individual compromises without resolving these targets (Fig. A3).

### 3.4 Consistent numerical evaluation of fixed formulations

Once a formulation is fixed, three questions remain: whether independently generated formulations of the same task give similar results, whether numerical search needs an agent, and whether saved evaluations can be reproduced. Five independently generated GR4J formulations, each run with three optimizer seeds on a common runner over 1990–1999, answer the first. Two of the five had selected the Kling–Gupta efficiency (KGE) [26] as their objective, and each of their seeded searches was repeated with NSE substituted, so that fifteen endpoints share one objective: nine from the original NSE specifications and six from the interventions (Table B10). Under the shared NSE objective, the five formulations yielded closely similar fit scores, with a total NSE range of $9.2 \times 10^{-4}$ and a largest within-formulation seed range of $2.3 \times 10^{-4}$ (Fig. 5b). Replacing KGE by NSE in the two formulations shifted their mean NSE by about 0.06, two orders of magnitude more than the spread among formulations, so the choice of objective is a formulation decision with consequences.

The paired comparison with VIC at Tangnaihai answers the second question. A fixed six-parameter formulation was run in ten blocks under two arms that shared the same five starting vectors and the same 150 calibration evaluations: DDS alone, with no language-model calls, and an agent proposing every subsequent parameter vector from the running diagnostics, in the manner of the parameter-update role [12]. Median validation NSE was 0.872 for fixed search and 0.875 for continuing proposals; the median paired difference, fixed minus continuing, was −0.003 with a 95% bootstrap interval of −0.007 to 0.002, whose lower bound lies above the non-inferiority margin of −0.02 (Fig. 5a,c; Table B11). Fixed search therefore met the non-inferiority criterion at a matched budget of calibration evaluations. Equal evaluation counts understate the difference in cost: ten continuing-arm attempts were interrupted by transport errors, rate limits or a host reboot and were excluded under the preregistered rules, whereas no fixed-search attempt failed.

Eleven replay packages answer the third question. Each re-evaluated one saved parameter vector, several of them default vectors, with the archived environment and compared the numerical outputs with the saved record; all eleven reproduced their comparison metrics, and three metric files, for the repeated GR4J formulation, the GR4J reference calibration and FSM2, were byte-identical (Table B12). Each replay reproduces one evaluation; full seeded searches and fresh agent generation were not replayed.

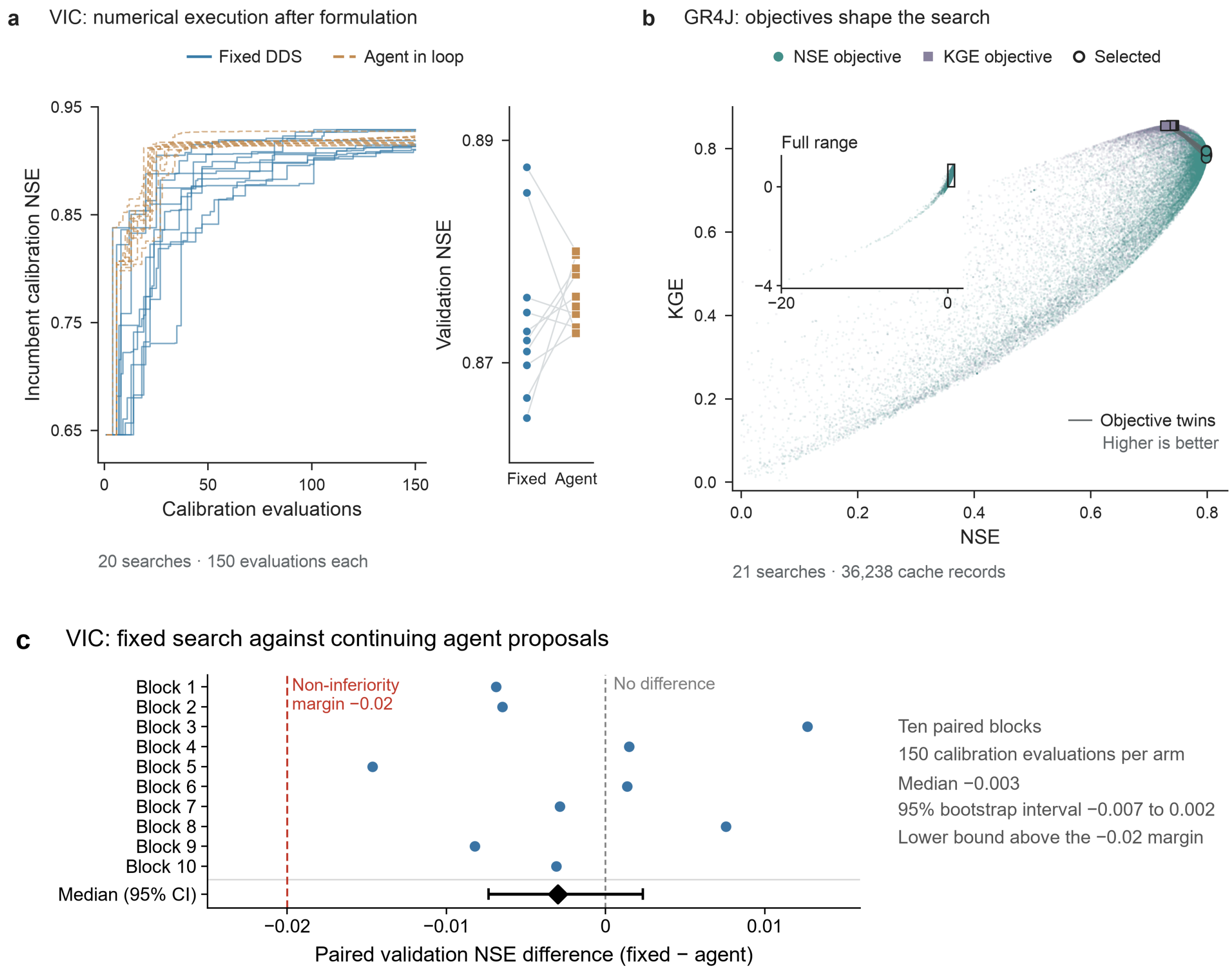

**Fig. 5 Fixed formulations under numerical search. a**, Ten paired VIC blocks at Tangnaihai: incumbent calibration NSE against evaluations for fixed DDS (blue, solid) and continuing agent proposals (orange, dashed), each with 150 calibration evaluations, and the paired validation NSE of the calibration-selected vectors (right). **b**, GR4J scores from 21 searches over 1990–1999 in NSE–KGE space: NSE-objective searches (green) and KGE-objective searches (purple), with selected endpoints circled and six segments linking matched KGE-to-NSE objective interventions; the inset shows the full range. **c**, Paired validation NSE differences, fixed minus continuing, for the ten blocks, with the median (diamond) and its 95% complete-block bootstrap interval (bar), the line of no difference (grey) and the non-inferiority margin of −0.02 (red).

Two auxiliary comparisons place these results. With a hand-written, literature-derived SAC-SMA setup at Ibex, mean validation NSE at a nominal budget of 3,000 evaluations was 0.53 for shuffled complex evolution (SCE-UA) [27], 0.50 for DDS and 0.43 for random search, although DDS reached the higher calibration fit (Fig. A4; Tables B13 and B14); the numerical method remains a consequential choice after formulation. Against established references, the generated GR4J workflow and the airGR reference calibration [28] both reached NSE 0.80 and KGE 0.79 at basin L0123001, with fitted parameters differing by 0.2–4.1%, using 1,000 and 234 evaluations respectively (Fig. A5). For SAC-SMA at the Blanco River, the generated standalone workflow reached validation NSE 0.70 against 0.63 for the published CAMELS benchmark [29], with KGE 0.69 against 0.70 and a larger positive volume bias (+24.8% against −0.5%) under a different forcing, snow and evapotranspiration configuration, and none of its eleven fitted parameters lay inside the range of ten reference parameter files; the supplied bound on upper-zone free-water storage (5–150 mm) excluded the reference values of 490–698 mm (Fig. A5; Table B15).

## 4 Discussion and conclusion

Applying a process-based model to a site and a question involves a set of choices that shape what the model is tested against: which parameters may vary and within what bounds, which simulated quantities are compared with which observations and in what units, which metric expresses the question and which criteria decide adequacy. The experiments reported here show that these choices can be written as an explicit, executable experiment definition, a contract and a runner, that one program checks, executes and assesses. The VIC contract illustrates the content: parameters drawn from the influence structure of the model, forcing corrections excluded as a matter of principle, a parameter dropped because the generated inputs cannot carry it, and ranges narrowed to the model's feasibility constraint. Each is a scientific decision about the application, and the definition states it once, where it can be read.

The snow-model experiment shows how model knowledge enters such a definition. Physical meaning reaches the software through names and units. The observation the agents intended was the snow water equivalent, but the runner exposes it under an interface name, and the timescales they chose were plausible in hours but were read as seconds, so the search explored albedo decay on a timescale of minutes while every evaluation remained numerically valid. The agents without the knowledge understood the snow; what they lacked were two facts about this model's interface, which the knowledge infrastructure supplies together with the parameter domains that the GR4J test showed to agree with the developers' experience. The SAC-SMA comparison shows the complementary case, in which the supplied knowledge was incomplete: the bound on upper-zone free-water storage excluded the values of a published reference configuration. Because the bound is recorded in the contract with its source, it can be audited and corrected once for every later use. The reference configuration differs in forcing, snow and evapotranspiration treatment, so the two parameter sets belong to different applications, and the record makes that difference inspectable.

Asking a model to match two observed quantities at once turns calibration into a test of the whole application, and the component decisions are then the information a modeller acts on. VIC matched discharge at Xixian while its evapotranspiration moved away from the satellite product, and WOFOST matched grain mass in the calibration years but not in the validation years while biomass stayed within its ceiling. Under the tested formulations these are unresolved targets, and the record names the variable and the criterion. What to examine next follows from the pair. For VIC, the questions are whether a monthly gridded evapotranspiration product constrains a catchment of about 10,000 $km^2$ at the resolution of the model, whether the active parameters govern the evaporative partition, and whether a discharge-dominated objective should trade against it; for WOFOST, whether five calibration years of destructive sampling identify seven parameters, and whether a fixed error ceiling is the right test for an interannual crop series. These are questions of observation representativeness, parameter domain, objective and process representation. The acceptance counts have the same scope: they apply the declared criteria of this task, including the fallback and baseline rules used where a convention stores no ceiling.

Fixing the definition before search is what makes comparison and reuse possible. Five GR4J formulations written in independent sessions reached the same fit once they shared an objective, and the objective itself moved the result by two orders of magnitude more, so the formulation decisions that matter are visible in the contract and can be compared across studies. Numerical search under a fixed definition met its non-inferiority criterion with no agent calls, and saved evaluations replay exactly, so a calibration can be re-executed by anyone holding the knowledge package, with seeded optimizers accounting for the only remaining randomness. The earlier study of language models inside the calibration loop [12] tested whether a model can propose useful parameter updates from diagnostics; the present study tests whether model knowledge can be turned into an experiment definition that numerical software executes. The paired comparison connects the two by showing what the definition delivers when the proposals are removed (Table B16).

Each model family was calibrated at one site, so the results establish formulation across model types and leave transfer across hydroclimates untested; the acceptance bands are those recorded in each convention, and fallback and baseline rules decided the components for which a convention stores no ceiling; the principal experiments used one language-model configuration, with three further configurations tested for single-objective authoring; and a knowledge package is complete only to the extent that its sources are, as the SAC-SMA bound shows. The same apparatus can test the questions these conditions raise: whether a frozen knowledge package and experiment definition hold at a new site, with a different observation combination or under a different task, and whether a full seeded search can be reproduced from the archive. The knowledge library from which the six packages were drawn describes 280 process-based models; the calibrations tested here are those in Table 1. What accumulates, if experiment definitions are written and kept in this form, is a record of model applications in which the scientific choices are explicit, can be compared across sites and studies, and can be re-executed when the knowledge improves.

## Data and code availability

The general calibration harness (GCH) is available at https://github.com/lzwei196/GCH. The knowledge infrastructure from which the model packages were drawn is available at https://github.com/lzwei196/KISS-Knowledge-Infrastructure-for-Scientific-Simulation. The models are open source: WOFOST through PCSE (https://github.com/ajwdewit/pcse), HBV through HBV-bmi (https://github.com/Daafip/HBV-bmi), SUMMA (https://github.com/CH-Earth/summa), VIC (https://github.com/UW-Hydro/VIC), MODFLOW 6

(https://github.com/MODFLOW-ORG/modflow6), CRHM (https://github.com/CentreForHydrology/CRHM), FSM2 (https://github.com/RichardEssery/FSM2), GR4J through airGR (https://cran.r-project.org/package=airGR), SAC-SMA (https://github.com/NOAA-OWP/sac-sma) and HYMOD as distributed with SPOTPY (https://github.com/thouska/spotpy). Numerical search used the optimizers of SPOTPY and pymoo (https://github.com/anyoptimization/pymoo). Observations are from the Kellogg Biological Station LTER (https://lter.kbs.msu.edu), the HYDAT archive of the Water Survey of Canada (https://wateroffice.ec.gc.ca), CanSWE (https://doi.org/10.5281/zenodo.4734371), the SNOTEL network (Mt Rose, https://wcc.sc.egov.usda.gov/nwcc/site?sitenum=652), GLEAM (https://www.gleam.eu), the Sagehen example distributed with MODFLOW 6 (https://github.com/MODFLOW-ORG/modflow6-examples) and the CAMELS dataset (https://ral.ucar.edu/solutions/products/camels). The Chinese discharge records for Xixian and Tangnaihai are subject to redistribution restrictions.

## Acknowledgements

We gratefully acknowledge funding provided by the National Natural Science Foundation of China (52279018, 52121006, and U2240203), National Basic Research Special Fund for Public Welfare Research Institutes (YDS25002), 2025 Innovation Fund of the National Key Laboratory of Water Disaster Prevention (Yk325002, 524015252, 5240152M2), Fundamental Research Funds for the Central Universities (B240201060), Open Research Fund of National Engineering Research Center of Water Resources Efficient Utilization and Engineering Safety (1524020003).

## Author contributions

Z.L. designed the study. Z.L., Z.Z. and Y.Z. led and performed the experiments, including the knowledge-guided formulations and multivariable calibrations across the six model families, the FSM2 and GR4J information-condition experiments, the repeated GR4J formulations, the paired VIC comparison of fixed search with continuing agent proposals, the reference comparisons and the numerical replays and optimizer comparisons. L.Z. and Y.L. reviewed and verified the experiments. J.J. and J.Z. supervised the work and acquired the funding.

## Competing interests

The authors declare no competing interests.

## References

1. Clark, M. P. et al. The evolution of process-based hydrologic models: historical challenges and the collective quest for physical realism. *Hydrol. Earth Syst. Sci.* **21**, 3427–3440 (2017).
2. Fatichi, S. et al. An overview of current applications, challenges, and future trends in distributed process-based models in hydrology. *J. Hydrol.* **537**, 45–60 (2016).
3. Addor, N. & Melsen, L. A. Legacy, rather than adequacy, drives the selection of hydrological models. *Water Resour. Res.* **55**, 378–390 (2019).

4. Beven, K. & Freer, J. Equifinality, data assimilation, and uncertainty estimation in mechanistic modelling of complex environmental systems using the GLUE methodology. *J. Hydrol.* **249**, 11–29 (2001).
5. Mai, J. Ten strategies towards successful calibration of environmental models. *J. Hydrol.* **620**, 129414 (2023).
6. Gupta, H. V., Wagener, T. & Liu, Y. Reconciling theory with observations: elements of a diagnostic approach to model evaluation. *Hydrol. Process.* **22**, 3802–3813 (2008).
7. Rakovec, O. et al. Multiscale and multivariate evaluation of water fluxes and states over European river basins. *J. Hydrometeorol.* **17**, 287–307 (2016).
8. Knoben, W. J. M., Freer, J. E. & Woods, R. A. Technical note: inherent benchmark or not? Comparing Nash–Sutcliffe and Kling–Gupta efficiency scores. *Hydrol. Earth Syst. Sci.* **23**, 4323–4331 (2019).
9. Efstratiadis, A. & Koutsoyiannis, D. One decade of multi-objective calibration approaches in hydrological modelling: a review. *Hydrol. Sci. J.* **55**, 58–78 (2010).
10. Moriasi, D. N., Gitau, M. W., Pai, N. & Daggupati, P. Hydrologic and water quality models: performance measures and evaluation criteria. *Trans. ASABE* **58**, 1763–1785 (2015).
11. Seidel, S. J., Palosuo, T., Thorburn, P. & Wallach, D. Towards improved calibration of crop models – where are we now and where should we go? *Eur. J. Agron.* **94**, 25–35 (2018).
12. Zhu, Z. et al. Large language models as calibration agents in hydrological modeling: feasibility and limitations. *Geophys. Res. Lett.* **53**, e2025GL120043 (2026).
13. Yan, S. et al. AI agent for hydrologic modeling: definition, development, and application. *Geophys. Res. Lett.* **53**, e2025GL119814 (2026).
14. Li, Z. et al. HydroAgent: closing the gap between frontier LLMs and human experts in hydrologic model calibration via simulator-grounded RL. Preprint at https://doi.org/10.48550/arXiv.2605.17792 (2026).
15. Yang, Q. et al. HydroAgent: formalizing forecaster expertise into skill-orchestrated flood forecasting workflows. Preprint at https://doi.org/10.48550/arXiv.2607.23983 (2026).
16. Jimenez, C. E. et al. SWE-bench: can language models resolve real-world GitHub issues? In *Proc. 12th International Conference on Learning Representations* (ICLR, 2024).
17. Geng, H. et al. ScienceIDE: turning world's scientific codebase into agent learnable environments. Preprint at https://doi.org/10.48550/arXiv.2609.19134 (2026).
18. Li, Z. et al. KISS—knowledge infrastructure for scientific simulation: a scaffolding for agentic Earth science. Preprint at https://doi.org/10.48550/arXiv.2605.17856 (2026).
19. Zhang, L. et al. Foundation models as assistive tools in hydrometeorology: opportunities, challenges, and perspectives. *Water Resour. Res.* **61**, e2024WR039553 (2025).
20. Nash, J. E. & Sutcliffe, J. V. River flow forecasting through conceptual models part I—a discussion of principles. *J. Hydrol.* **10**, 282–290 (1970).
21. Tolson, B. A. & Shoemaker, C. A. Dynamically dimensioned search algorithm for computationally efficient watershed model calibration. *Water Resour. Res.* **43**, W01413 (2007).
22. Vrugt, J. A. Markov chain Monte Carlo simulation using the DREAM software package: theory, concepts, and MATLAB implementation. *Environ. Model. Softw.* **75**, 273–316 (2016).
23. Deb, K., Pratap, A., Agarwal, S. & Meyarivan, T. A fast and elitist multiobjective genetic algorithm: NSGA-II. *IEEE Trans. Evol. Comput.* **6**, 182–197 (2002).
24. Essery, R., Mazzotti, G., Barr, S., Jonas, T., Quaife, T. & Rutter, N. A Flexible Snow Model (FSM 2.1.1) including a forest canopy. *Geosci. Model Dev.* **18**, 3583–3605 (2025).
25. Perrin, C., Michel, C. & Andréassian, V. Improvement of a parsimonious model for streamflow simulation. *J. Hydrol.* **279**, 275–289 (2003).
26. Gupta, H. V., Kling, H., Yilmaz, K. K. & Martinez, G. F. Decomposition of the mean squared error and NSE performance criteria: implications for improving hydrological modelling. *J. Hydrol.* **377**, 80–91 (2009).

27. Duan, Q., Sorooshian, S. & Gupta, V. Effective and efficient global optimization for conceptual rainfall-runoff models. *Water Resour. Res.* **28**, 1015–1031 (1992).

28. Coron, L., Thirel, G., Delaigue, O., Perrin, C. & Andréassian, V. The suite of lumped GR hydrological models in an R package. *Environ. Model. Softw.* **94**, 166–171 (2017).

29. Newman, A. J. et al. Development of a large-sample watershed-scale hydrometeorological data set for the contiguous USA: data set characteristics and assessment of regional variability in hydrologic model performance. *Hydrol. Earth Syst. Sci.* **19**, 209–223 (2015).

30. Houska, T., Kraft, P., Chamorro-Chavez, A. & Breuer, L. SPOTting model parameters using a ready-made Python package. *PLoS ONE* **10**, e0145180 (2015).

31. Blank, J. & Deb, K. pymoo: multi-objective optimization in Python. *IEEE Access* **8**, 89497–89509 (2020).

32. Soothar, R. K. et al. Effect of different soil moisture regimes on plant growth and water use efficiency of sunflower: experimental study and modeling. *Bull. Natl. Res. Cent.* **45**, 121 (2021).

# Appendix A: Additional figures

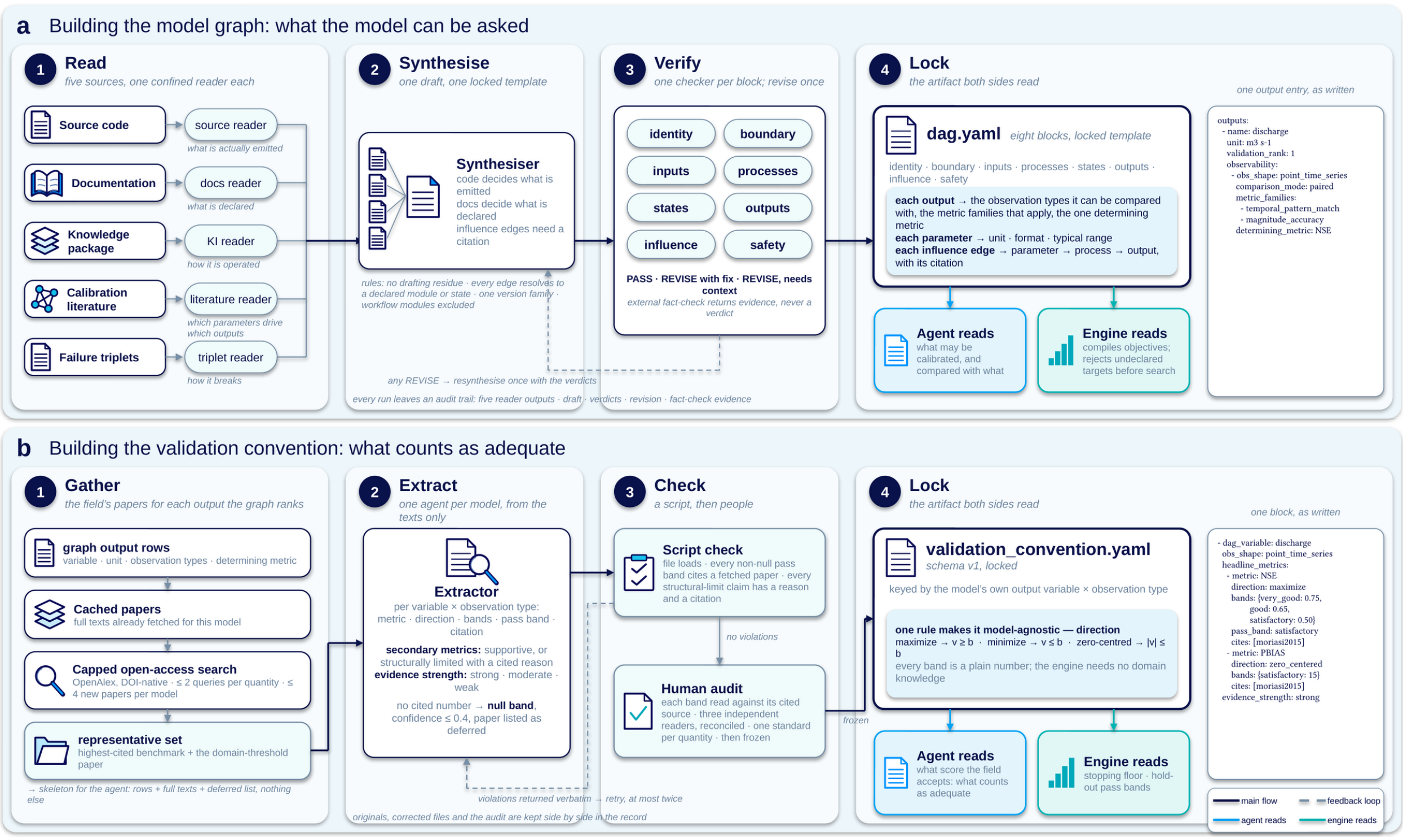

**Fig. A1 Construction of the model graph and the validation convention.** **a**, The model graph. Five readers each read one source (source code, documentation, the operating package, the calibration literature and recorded failure triplets) and emit structured facts; a synthesiser merges them into a locked eight-block template; one verifier per block returns a pass or revise verdict, after which the synthesiser runs at most once more. The locked file declares, for each output, the observation types it can be compared with, the applicable metric families and the determining metric; the agent reads it to formulate and the engine reads it to compile objectives. **b**, The validation convention. For each ranked output, papers are gathered from cached texts and a capped open-access search; an extraction agent records the metric, its direction, the numeric bands and the citation for each variable and observation type; a script checks that every band cites a fetched paper; three readers audit each band against its source before the file is frozen. The agent reads the convention to learn what counts as adequate; the engine reads it for the acceptance bands. Example entries at the right show one output block and one convention block as written.

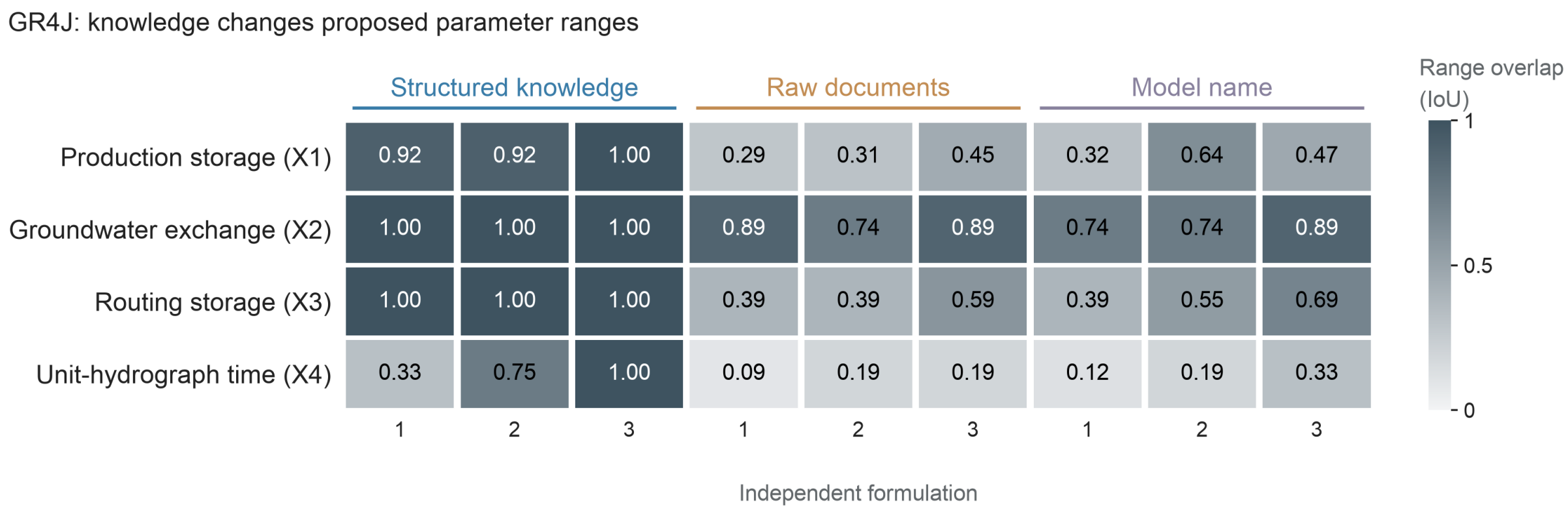

**Fig. A2 Knowledge changes the GR4J parameter ranges that agents propose.** Intersection over union of each specification's transformed parameter range with the empirical 10th–90th percentile box of the model developers [25], for nine independent GR4J specifications: three written with the knowledge infrastructure, three with raw documentation and three with the model name only. Darker cells indicate larger overlap. The ranges were assessed without calibration. All three model-name specifications selected the target Q rather than the exposed runner variable Qsim.

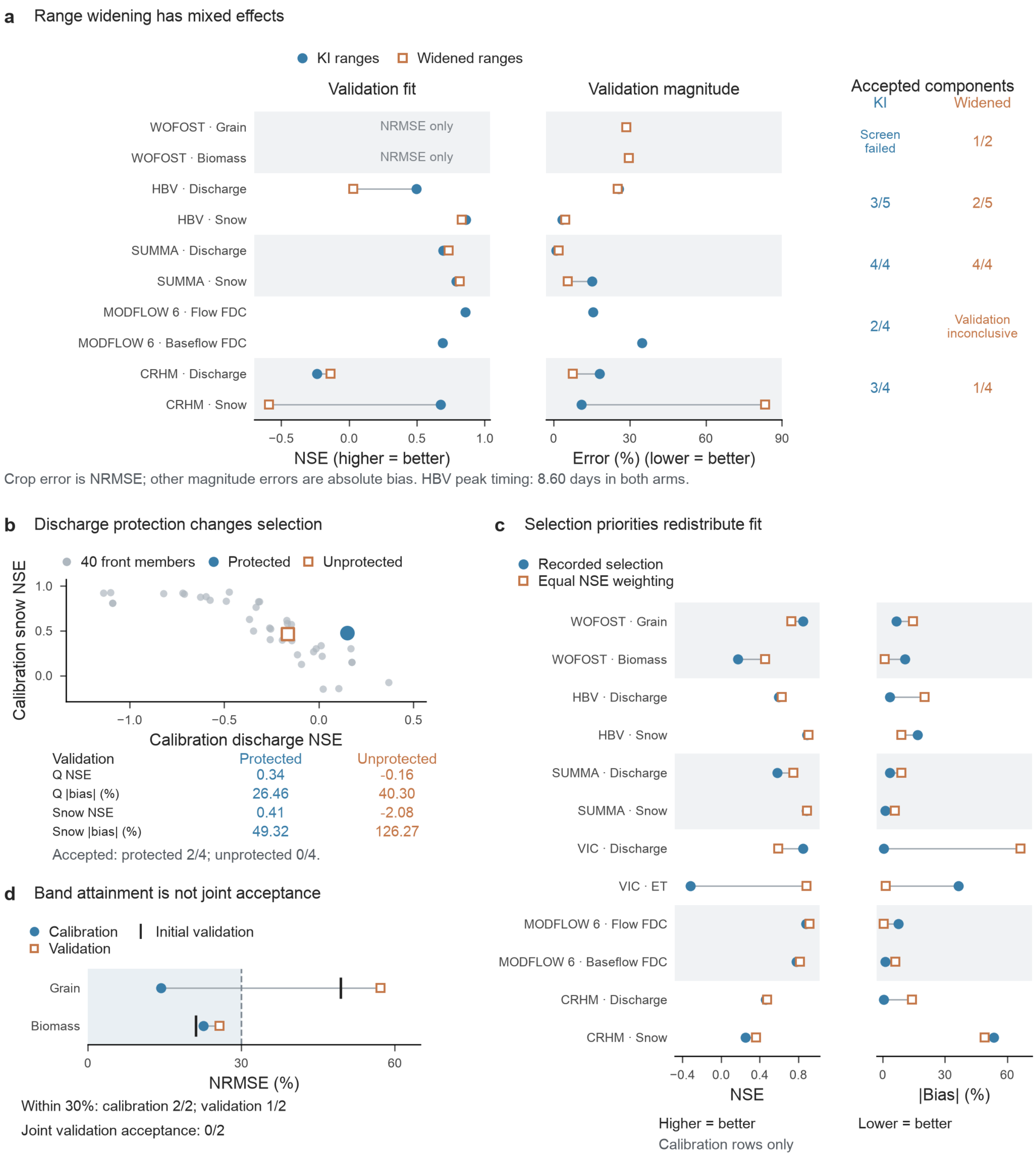

**Fig. A3 Parameter ranges, selection rules and assessment criteria change calibration compromises. a**, Validation outcomes of five matched parameter-range diagnostics comparing the supplied knowledge-infrastructure ranges (blue circles) with ranges widened threefold before transformation and non-negativity constraints (orange squares), one seed per arm at a declared budget of 150. Eight arms have assessable validation results; the WOFOST supplied-range sensitivity screen returned no finite losses, and the widened MODFLOW 6 arm completed optimization but its split probe left validation inconclusive. Crop error is NRMSE and other magnitude errors are absolute PBIAS; accepted-component counts use the diagnostic assessment rules, and HBV includes a fifth, timing component. **b**, All 40 members of one CRHM front after 120 evaluations projected onto calibration discharge and snow NSE, with the discharge-protected (blue) and unprotected (orange) selections and their validation metrics listed below. **c**, For all twelve targets of the six-family application, the recorded

selection (blue) against the cached candidate that minimizes the mean of max(0, 1 − NSE) across targets (orange), both scored on calibration data. **d**, WOFOST NRMSE of the selected member in calibration (blue) and validation (orange) against the uncalibrated validation reference (bar); the shaded band marks the 30% ceiling. Q, discharge; ET, evapotranspiration; FDC, flow-duration curve.

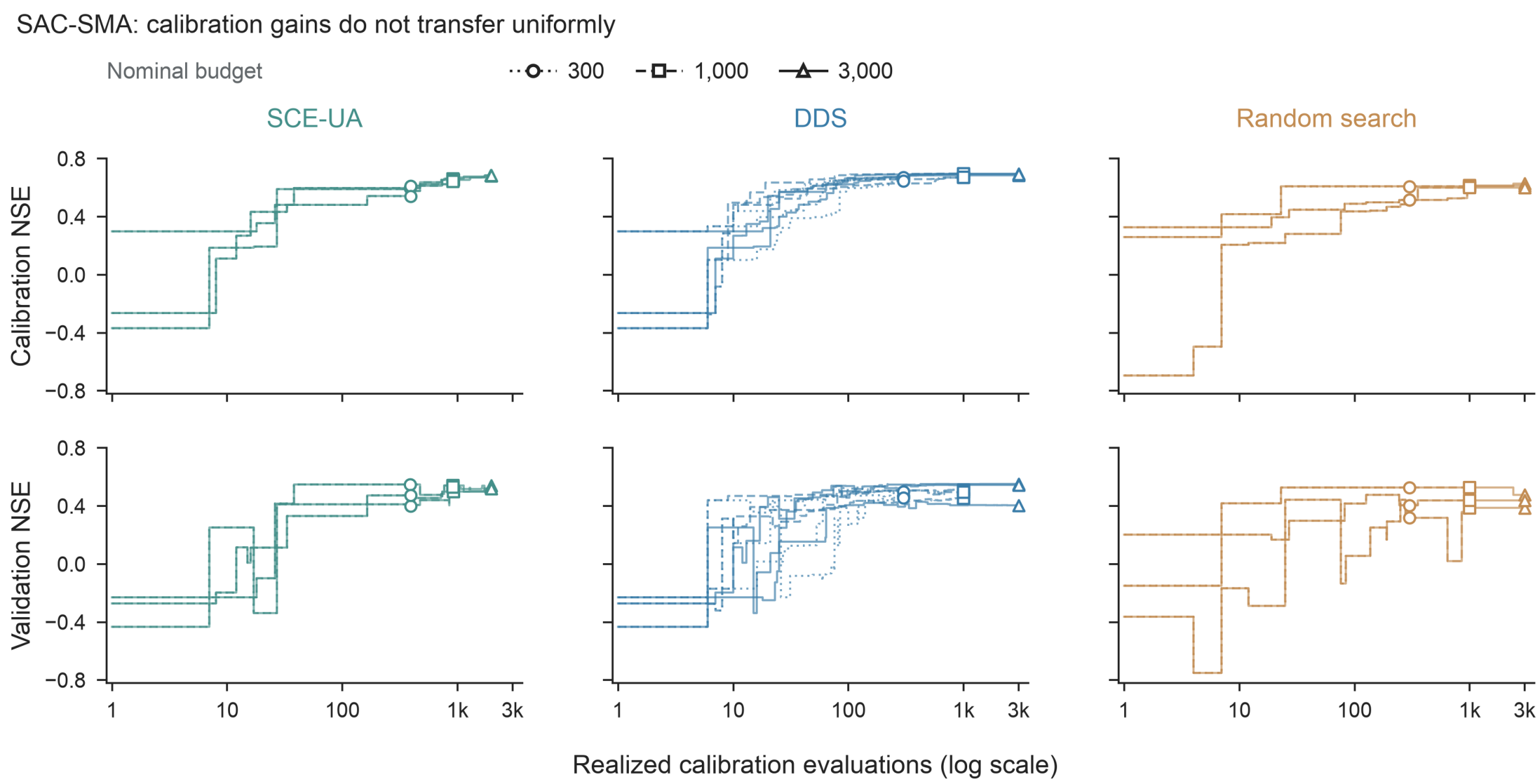

**Fig. A4 Optimizer comparison under a fixed SAC-SMA formulation at Ibex.** Twenty-seven searches with SCE-UA, DDS and uniform random search at nominal budgets of 300, 1,000 and 3,000 evaluations and three seeds; trajectories show calibration (top) and validation (bottom) NSE of the calibration-selected incumbent against realized evaluations and end at the realized count. The setup is literature-derived and was prepared without knowledge-guided authoring; it isolates the effect of the numerical method once a formulation is fixed (Appendix G; Tables B13 and B14).

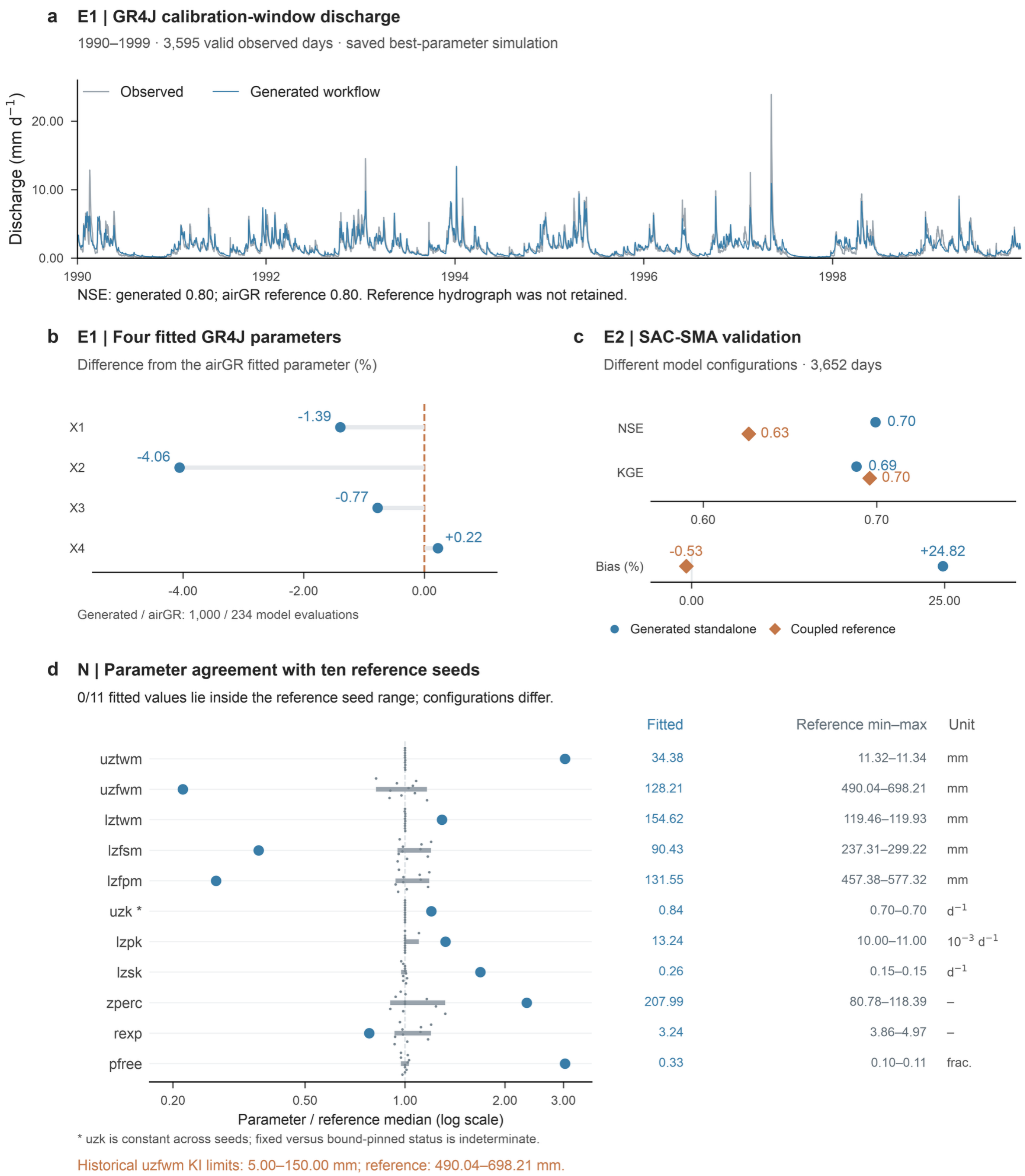

**Fig. A5 Generated calibrations against established references. a**, Observed discharge (grey) and the generated GR4J workflow simulation (blue) at basin L0123001 over the 1990–1999 calibration window (3,595 valid days). **b**, Percentage difference of the four fitted GR4J parameters from the airGR reference estimates. **c**, SAC-SMA validation at the Blanco River (October 1989 to September 1999): NSE, KGE and percentage bias of the generated standalone workflow (blue circles) and the published coupled benchmark (orange diamonds). **d**, Eleven fitted SAC-SMA parameters (blue) relative to the median of ten reference parameter files (grey points and ranges), with fitted values, reference ranges and units listed at the right; the asterisk marks a parameter that is constant across the reference files, and the footnote gives the supplied bounds on upper-zone free-water storage.

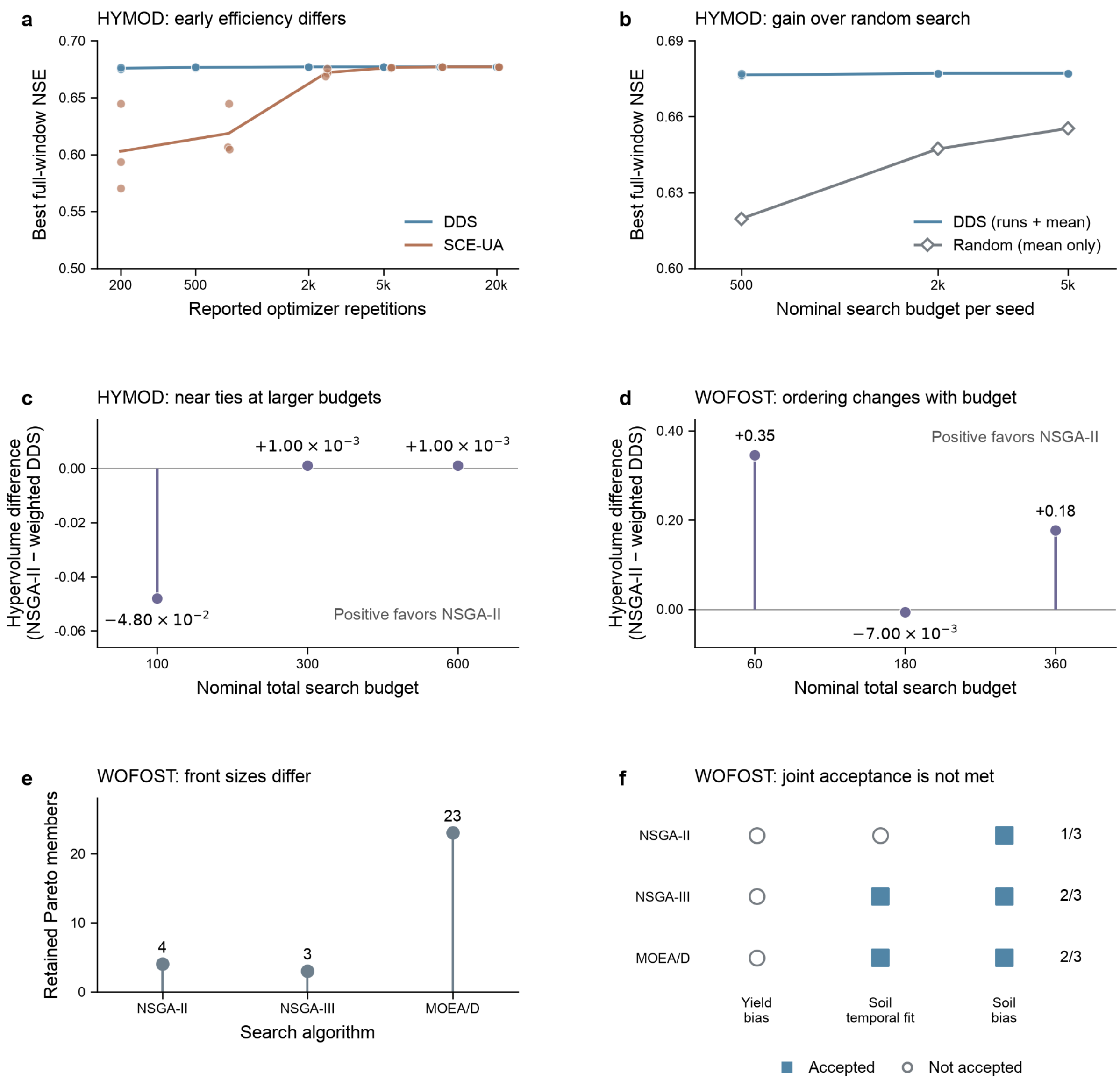

**Fig. A6 Numerical-search comparisons under fixed formulations.** **a**, Best full-window NSE for HYMOD at Leaf River under DDS (blue) and SCE-UA (orange) at six nominal budgets with seeds 0–2; dots are the 36 endpoints and lines connect the three-seed means at the mean reported optimizer repetitions. **b**, HYMOD DDS endpoints and means against random-search means at three budgets; random-search results are available only as three-seed means. **c**,**d**, Differences in hypervolume between NSGA-II and the pooled candidates of five weighted DDS searches under equal nominal total budgets for HYMOD (**c**) and WOFOST (**d**); positive values favour NSGA-II. **e**, Retained Pareto-front sizes for three WOFOST multiobjective backends at a nominal budget of 40. **f**, Component-acceptance decisions for the same three backends; filled squares, accepted; open circles, not accepted; counts at the right give accepted components out of three.

## Appendix B: Additional tables

Values are displayed to two decimals, with scientific notation for small differences. NSE, Nash–Sutcliffe efficiency; KGE, Kling–Gupta efficiency; PBIAS, percentage bias; NRMSE, root-mean-square error normalized by the mean observation.

**Table B1 Components of a knowledge infrastructure and who reads them.** The packages used in this study share the same nine components. The three procedural components are read by the agent; the model graph and the validation convention are declarative and are read by the agent and the harness.

| Component | File(s) | Content | Read by |
|---|---|---|---|
| Operating manual | SKILL.md | What the model simulates, runtime, pipeline overview, parameter reference, unit conventions and a map of which file to read when | Agent |
| Staged protocols | docs/sN_stage.md (s0–s9) | One procedure per pipeline stage (forcing, configuration, execution, parsing, validation, calibration) with verification steps and known traps | Agent |
| Validated operators | tools/ | One executable per stage plus calib_run.py, the runner that a specification invokes | Agent (composes); harness (executes) |
| Diagnostic triplets | diagnostics/triplets.yaml | Symptom, diagnosis and remedy for the model's recorded failure modes | Agent |
| Model graph | dag.yaml | Scope of the model; every input, parameter and output with units, formats and typical ranges; for each output the comparable observation types and the applicable metric families | Harness (objectives); agent |
| Validation convention | docs/validation_convention.yaml | Per-variable metrics, directions and pass bands with the studies they are taken from | Harness (assessment); agent |
| Input–output specification | docs/format_spec.yaml | Exact input and output shapes, projected from the graph and the triplets | Agent (runner authoring) |
| Literature index | docs/gathered_papers.json; docs/papers_index.md | The studies from which the model's conventions are drawn | Agent |
| Manifest | knowledge_infrastructure.yaml | Package summary, validation tier and the measured skill of the reference run | Agent; harness |

**Table B2 Experimental designs and evaluated capabilities.** Shared samples and derived diagnostics are identified in Appendices C–G; the individual numerical observations appear in the tables cited in the last column.

| Question group | Experiment | Design | Evaluated outcomes | Supporting material |
|---|---|---|---|---|
| Formulation across models | Six-family calibration | 6 model-specific formulations; 23 assessed components; different interfaces and observation structures | Task-specific preparation, completion and multivariable scientific acceptance | Appendix D; Fig. 4; Tables B7–B9 |
| Formulation across models | Single-objective phase and cross-vendor authoring | 6 single-objective formulations executed once; 3 poor-baseline retests; 15 authorings by 4 model configurations; 12 blinded decisions by 3 vendors | Generation and execution without revision, improvement where defaults were poor, authoring cost, stability of optimizer choice across models | Appendix D; Tables B3–B5 |
| Formulation across models | Workflow-generation examples and direct drivers | 2 generated workflows; 1 scalar-objective probe; 4 hand-written multivariable applications | Generated code and scientific choices; engine support for multivariable tasks | Appendix D; Tables B17 and B18 |
| Knowledge contribution | FSM2 knowledge ablation | 3 information conditions × 5 fresh specifications; 8 initial and 13 repair searches; runner supplied | Initial execution, validation performance, target and unit corrections | Appendix E; Fig. 2; Table B6 |
| Knowledge contribution | GR4J range and target test | 3 information conditions × 3 fresh specifications; static assessment | Empirical range overlap and observation-target mapping | Appendix E; Fig. A2 |
| Reference comparison | Reference calibration and parameter comparisons | 1 GR4J and 1 SAC-SMA calibration; 10 SAC-SMA reference seed files | Reference fit, validation metrics, parameter identities and values | Appendix E; Fig. A5; Table B15 |
| Assessment diagnostics | Range and selection diagnostics | 10 range arms in 5 families; 12 cached target comparisons; 2 rules on one CRHM front | Effects of constraints, scientific priorities and assessment criteria | Appendix G; Fig. A3; Tables B22 and B23 |
| Consistency and numerical reproducibility | Repeated GR4J formulation | 5 specifications × 3 seeds plus 6 objective interventions; 21 searches | Consistency under a common objective and effects of objective choice | Appendix F; Table B10 |
| Consistency and numerical reproducibility | VIC fixed search versus continuing proposals | 10 paired blocks; 150 calibration evaluations per arm; shared starting points | Validation non-inferiority and optimization-stage agent calls | Appendix F; Table B11 |
| Consistency and numerical reproducibility | Numerical replay | 11 packages; one specified parameter evaluation per package | Agreement with original numerical metrics | Appendix F; Table B12 |
| Numerical diagnostics | Optimizer and multiobjective comparisons | 27 SAC-SMA searches; 36 HYMOD endpoints; scalar/multiobjective and backend diagnostics | Search performance, budget effects and multivariable compromises | Appendix G; Fig. A6; Tables B13, B14, B19–B21 and B24 |

**Table B3 Single-objective phase across six model families.** One agent-authored single-objective specification per family, executed by the common harness at 80 evaluations (author claude-fable-5, 22–23 August 2026). Stopped denotes the pre-check outcome that the defaults already met the declared target. Poor-baseline retests used DDS at 120 evaluations (HBV, SUMMA) or 600 evaluations with three seeds (CRHM, Belly River). Values are transcribed from the coordinator session record.

| Model | Target variable | Optimizer selected | Outcome at 80 evaluations | Time per evaluation (s) | Poor-baseline retest, holdout NSE before and after |
|---|---|---|---|---|---|
| WOFOST | Grain mass | DDS | Completed, promotable | 6.9 | — |
| HBV | Discharge | DDS | Stopped, default adequate | — | 0.59 → 0.73 |
| SUMMA | Discharge | DDS | Stopped, default adequate | — | 0.40 → 0.61 and 0.70 (two seeds) |
| VIC | Discharge | DDS | Completed, promotable | 1,074 | — |
| MODFLOW 6 | Streamflow | DDS | Completed, promotable | 17.1 | — |
| CRHM | Discharge | SCE-UA | Stopped, default adequate | — | 0.44 → 0.71–0.73 (DDS); 0.67–0.71 (SCE-UA) |

**Table B4 Authoring cost and cross-vendor authoring of single-objective workflows.** Output tokens and cost are from the command-line usage record; GPT-5.5 usage was not captured. Every authoring produced a runnable specification; all selected DDS except the claude-fable-5 CRHM authoring, which selected SCE-UA.

| Author | Model | Output tokens | Cost (USD) | Authoring time (min) |
|---|---|---|---|---|
| claude-fable-5 | WOFOST | 67,395 | 11.24 | 17 |
| claude-fable-5 | HBV | 72,086 | 12.50 | 30 |
| claude-fable-5 | VIC | 29,913 | 8.98 | 23 |
| claude-fable-5 | MODFLOW 6 | 46,561 | 9.79 | 23 |
| claude-fable-5 | SUMMA | 16,575 | 3.57 | 11 |
| claude-fable-5 | CRHM | 87,042 | 14.21 | 34 |
| claude-sonnet-5 | WOFOST | 78,528 | 4.38 | 29 |
| claude-sonnet-5 | HBV | 59,665 | 3.37 | 24 |
| claude-sonnet-5 | MODFLOW 6 | 9,008 | 2.65 | 14 |
| claude-opus-4-8 | WOFOST | 21,963 | 3.01 | 11 |
| claude-opus-4-8 | HBV | 46,287 | 4.47 | 16 |
| claude-opus-4-8 | MODFLOW 6 | 37,893 | 3.89 | 12 |
| GPT-5.5 | WOFOST | — | — | 13 |
| GPT-5.5 | HBV | — | — | 9 |
| GPT-5.5 | MODFLOW 6 | — | — | 9 |

**Table B5 Blinded optimizer choices across three vendors.** Single-shot prompts (24 August 2026) offered DDS, SCE-UA, DREAM and NSGA-II with one-line descriptions; each framing described one calibration request without naming a model. The expected answers follow the objective structure: point estimate, DDS;

uncertainty, DREAM; two variables with a trade-off, NSGA-II. For the rugged-surface framing the textbook answer is SCE-UA; all three vendors chose DDS on budget grounds.

| **Framing** | **Claude** | **GPT** | **Kimi** |
|---|---|---|---|
| Single variable, point estimate | DDS | DDS | DDS |
| Single variable, parameter uncertainty | DREAM | DREAM | DREAM |
| Two observed variables, explicit trade-off | NSGA-II | NSGA-II | NSGA-II |
| Rugged surface, moderate budget | DDS | DDS | DDS |

**Table B6 FSM2 initial and repaired searches.** K, knowledge infrastructure; R, raw documentation; M, model name and task schema. Repairs reuse the initial specifications. Two M specifications contain the unsupported parameter bstb and fall outside the prescribed target and unit repair scope. Constraints and caching account for the difference between proposals and completed executions.

| Run | Preparation stage | Objective | Proposals | Completed model executions | Calibration NSE | Validation NSE | Validation KGE | Validation PBIAS (%) |
|---|---|---|---|---|---|---|---|---|
| K1_seed0 | Initial | NSE | 300 | 260 | 0.91 | 0.92 | 0.80 | −17.42 |
| K2_seed0 | Initial | KGE | 200 | 186 | 0.83 | 0.81 | 0.63 | −31.35 |
| K3_seed0 | Initial | NSE | 250 | 229 | 0.92 | 0.94 | 0.82 | −15.62 |
| K4_seed0 | Initial | NSE | 250 | 218 | 0.90 | 0.91 | 0.75 | −21.33 |
| K5_seed0 | Initial | NSE | 250 | 224 | 0.92 | 0.94 | 0.83 | −15.09 |
| M1_seed0_r4target | Target correction | KGE | 300 | 273 | 0.70 | 0.66 | 0.49 | −42.27 |
| M1_seed0_r4units | Unit correction | KGE | 300 | 268 | 0.92 | 0.93 | 0.76 | −19.97 |
| M3_seed0_r4target | Target correction | KGE | 300 | 277 | 0.68 | 0.62 | 0.45 | −45.01 |
| M3_seed0_r4units | Unit correction | KGE | 300 | 271 | 0.93 | 0.94 | 0.78 | −18.55 |
| M4_seed0_r4target | Target correction | KGE | 500 | 431 | 0.75 | 0.75 | 0.59 | −33.81 |
| M4_seed0_r4units | Unit correction | KGE | 500 | 428 | 0.95 | 0.99 | 0.91 | −7.81 |
| R1_seed0 | Initial | KGE | 250 | 230 | 0.68 | 0.63 | 0.45 | −44.79 |
| R1_seed0_r4units | Unit correction | KGE | 250 | 230 | 0.95 | 0.98 | 0.87 | −10.49 |
| R2_seed0 | Initial | KGE | 300 | 261 | 0.72 | 0.69 | 0.51 | −40.33 |
| R2_seed0_r4units | Unit correction | KGE | 300 | 261 | 0.94 | 0.98 | 0.86 | −11.55 |
| R3_seed0_r4target | Target correction | KGE | 250 | 236 | 0.67 | 0.61 | 0.44 | −45.82 |
| R3_seed0_r4units | Unit correction | KGE | 250 | 217 | 0.94 | 0.96 | 0.82 | −15.44 |
| R4_seed0_r4target | Target correction | KGE | 250 | 226 | 0.68 | 0.62 | 0.45 | −45.10 |
| R4_seed0_r4units | Unit correction | KGE | 250 | 220 | 0.94 | 0.98 | 0.87 | −11.02 |
| R5_seed0 | Initial | KGE | 400 | 350 | 0.68 | 0.62 | 0.45 | −45.12 |
| R5_seed0_r4units | Unit correction | KGE | 400 | 357 | 0.95 | 0.98 | 0.86 | −11.54 |

**Table B7 Commissioning scientific metrics.** Commissioning used budgets of 40–200 evaluations. WOFOST commissioning used absolute PBIAS and a maximum-biomass operator, whereas Table B8 uses NRMSE and final-maturity biomass; the two are distinct formulations rather than a controlled budget comparison. WOFOST values originate from contemporaneous tool output; other values come from case reports.

| Model | Component | Commissioning metric | Commissioning validation |
|---|---|---|---|
| WOFOST | TWSO: magnitude | Absolute PBIAS (%) | 11.06 |
| WOFOST | TAGP: magnitude | Absolute PBIAS (%) | 4.39 |
| HBV | Q: temporal fit | NSE | 0.50 |
| HBV | Q: magnitude | Absolute PBIAS (%) | 25.77 |
| HBV | SWE: magnitude | Absolute PBIAS (%) | 3.41 |
| HBV | SWE: temporal fit | NSE | 0.86 |
| HBV | SWE: timing | Absolute mean peak-date bias (days) | 8.60 |
| SUMMA | Q: temporal fit | NSE | 0.60 |
| SUMMA | Q: magnitude | Absolute PBIAS (%) | 4.70 |
| SUMMA | SWE: temporal fit | NSE | 0.82 |
| SUMMA | SWE: magnitude | Absolute PBIAS (%) | 5.31 |
| VIC | Q: temporal fit | NSE | 0.75 |
| VIC | Q: magnitude | Absolute PBIAS (%) | 19.07 |
| VIC | ET: temporal fit | NSE | 0.09 |
| VIC | ET: magnitude | Absolute PBIAS (%) | 22.58 |
| MODFLOW 6 | Flow FDC: temporal fit | NSE | 0.88 |
| MODFLOW 6 | Flow FDC: magnitude | Absolute PBIAS (%) | 20.44 |
| MODFLOW 6 | Baseflow FDC: temporal fit | NSE | 0.70 |
| MODFLOW 6 | Baseflow FDC: magnitude | Absolute PBIAS (%) | 36.76 |
| CRHM | Q: magnitude | Absolute PBIAS (%) | 26.46 |
| CRHM | Q: temporal fit | NSE | 0.34 |
| CRHM | SWE: magnitude | Absolute PBIAS (%) | 49.32 |
| CRHM | SWE: temporal fit | NSE | 0.41 |

**Table B8 Scientific metrics of the six multivariable calibrations.** Declared-budget results. Q, discharge; SWE, snow water equivalent; ET, evapotranspiration; FDC, flow-duration curve; TWSO, grain mass; TAGP, aboveground biomass. Validation informed final front-member selection. Commissioning values appear separately in Table B7.

| Model | Component | Metric | Baseline validation | Calibration | Validation |
|---|---|---|---|---|---|
| WOFOST | TWSO: magnitude | NRMSE (%) | 49.38 | 14.26 | 57.24 |
| WOFOST | TAGP: magnitude | NRMSE (%) | 21.12 | 22.59 | 25.77 |
| HBV | Q: temporal fit | NSE | 0.43 | 0.60 | 0.54 |
| HBV | Q: magnitude | Absolute PBIAS (%) | 28.58 | 3.54 | 13.36 |
| HBV | SWE: magnitude | Absolute PBIAS (%) | 12.15 | 16.72 | 11.43 |
| HBV | SWE: temporal fit | NSE | 0.90 | 0.89 | 0.89 |
| HBV | SWE: timing | Absolute mean peak-date bias (days) | 7.60 | 8.20 | 7.60 |
| SUMMA | Q: temporal fit | NSE | 0.65 | 0.58 | 0.64 |

| Model | Component | Metric | Baseline validation | Calibration | Validation |
|---|---|---|---|---|---|
| SUMMA | Q: magnitude | Absolute PBIAS (%) | 5.36 | 3.46 | 2.82 |
| SUMMA | SWE: temporal fit | NSE | 0.77 | 0.88 | 0.80 |
| SUMMA | SWE: magnitude | Absolute PBIAS (%) | 18.80 | 1.17 | 11.29 |
| VIC | Q: temporal fit | NSE | 0.74 | 0.85 | 0.74 |
| VIC | Q: magnitude | Absolute PBIAS (%) | 10.78 | 0.40 | 4.13 |
| VIC | ET: temporal fit | NSE | 0.05 | −0.32 | −0.29 |
| VIC | ET: magnitude | Absolute PBIAS (%) | 25.42 | 36.44 | 31.02 |
| MODFLOW 6 | Flow FDC: temporal fit | NSE | 0.86 | 0.88 | 0.88 |
| MODFLOW 6 | Flow FDC: magnitude | Absolute PBIAS (%) | 34.20 | 7.62 | 4.09 |
| MODFLOW 6 | Baseflow FDC: temporal fit | NSE | 0.61 | 0.78 | 0.73 |
| MODFLOW 6 | Baseflow FDC: magnitude | Absolute PBIAS (%) | 56.14 | 1.29 | 12.03 |
| CRHM | Q: magnitude | Absolute PBIAS (%) | 47.97 | 0.57 | 10.63 |
| CRHM | Q: temporal fit | NSE | 0.45 | 0.46 | 0.56 |
| CRHM | SWE: magnitude | Absolute PBIAS (%) | 74.34 | 53.46 | 58.48 |
| CRHM | SWE: temporal fit | NSE | −0.27 | 0.25 | 0.18 |

**Table B9 Threshold attainment and joint component acceptance.** Inclusive loss ceilings are expressed in the metric units of Table B8. Joint acceptance additionally requires finite losses, temporal correlation where applicable and no deterioration from the validation baseline. Unavailable fields are not inferred; the fallback rules are described in the Methods.

| Model | Component | Stored ceiling | Ceiling met | Fallback check | No baseline deterioration | Joint accepted |
|---|---|---|---|---|---|---|
| WOFOST | TWSO: magnitude | ≤ 30.00% | No | None | No | No |
| WOFOST | TAGP: magnitude | ≤ 30.00% | Yes | None | No | No |
| HBV | Q: temporal fit | Not stored | Not recorded | None | Yes | Yes |
| HBV | Q: magnitude | ≤ 15.00% | Yes | None | Yes | Yes |
| HBV | SWE: magnitude | ≤ 15.00% | Yes | None | Yes | Yes |
| HBV | SWE: temporal fit | Not stored | Not recorded | None | No | No |
| HBV | SWE: timing | Not stored | Not recorded | None | Yes | Yes |
| SUMMA | Q: temporal fit | Not stored | Not recorded | None | No | No |
| SUMMA | Q: magnitude | ≤ 15.00% | Yes | None | Yes | Yes |
| SUMMA | SWE: temporal fit | Not stored | Not recorded | None | Yes | Yes |
| SUMMA | SWE: magnitude | ≤ 15.00% | Yes | None | Yes | Yes |
| VIC | Q: temporal fit | NSE ≥ 0.50 | Yes | None | Yes | Yes |
| VIC | Q: magnitude | Not stored | Not recorded | ≤ 30.00% | Yes | Yes |
| VIC | ET: temporal fit | NSE ≥ 0.50 | No | None | No | No |
| VIC | ET: magnitude | Not stored | Not recorded | ≤ 30.00% | No | No |

| Model | Component | Stored ceiling | Ceiling met | Fallback check | No baseline deterioration | Joint accepted |
|---|---|---|---|---|---|---|
| MODFLOW 6 | Flow FDC: temporal fit | NSE ≥ 0.50 | Yes | None | Yes | Yes |
| MODFLOW 6 | Flow FDC: magnitude | ≤ 15.00% | Yes | None | Yes | Yes |
| MODFLOW 6 | Baseflow FDC: temporal fit | NSE ≥ 0.50 | Yes | None | Yes | Yes |
| MODFLOW 6 | Baseflow FDC: magnitude | ≤ 15.00% | Yes | None | Yes | Yes |
| CRHM | Q: magnitude | ≤ 15.00% | Yes | None | Yes | Yes |
| CRHM | Q: temporal fit | Not stored | Not recorded | None | Yes | Yes |
| CRHM | SWE: magnitude | ≤ 15.00% | No | None | Yes | No |
| CRHM | SWE: temporal fit | Not stored | Not recorded | None | Yes | Yes |

**Table B10 GR4J repeated-formulation searches.** All 21 searches use the optimization window 1990–1999. The common-NSE analysis includes the nine original NSE searches and the six objective interventions of the two KGE specifications. Five specifications each contribute three numerical seeds.

| Run | Objective | Algorithm | Reported evaluations | Cap | NSE | KGE |
|---|---|---|---|---|---|---|
| C1_seed0 | KGE | SCE-UA | 2121 | 3000 | 0.74 | 0.86 |
| C1_seed1 | KGE | SCE-UA | 2135 | 3000 | 0.74 | 0.86 |
| C1_seed2 | KGE | SCE-UA | 1932 | 3000 | 0.74 | 0.86 |
| C1_seed0_nse | NSE | SCE-UA | 1935 | 3000 | 0.80 | 0.79 |
| C1_seed1_nse | NSE | SCE-UA | 1976 | 3000 | 0.80 | 0.79 |
| C1_seed2_nse | NSE | SCE-UA | 1921 | 3000 | 0.80 | 0.78 |
| C2_seed0 | NSE | SCE-UA | 1926 | 3000 | 0.80 | 0.78 |
| C2_seed1 | NSE | SCE-UA | 1921 | 3000 | 0.80 | 0.78 |
| C2_seed2 | NSE | SCE-UA | 1947 | 3000 | 0.80 | 0.79 |
| C3_seed0 | NSE | SCE-UA | 2113 | 3000 | 0.80 | 0.79 |
| C3_seed1 | NSE | SCE-UA | 1926 | 3000 | 0.80 | 0.78 |
| C3_seed2 | NSE | SCE-UA | 1930 | 3000 | 0.80 | 0.79 |
| C4_seed0 | NSE | DDS | 1500 | 1500 | 0.80 | 0.79 |
| C4_seed1 | NSE | DDS | 1500 | 1500 | 0.80 | 0.79 |
| C4_seed2 | NSE | DDS | 1500 | 1500 | 0.80 | 0.79 |
| C5_seed0 | KGE | SCE-UA | 1334 | 2000 | 0.74 | 0.86 |
| C5_seed1 | KGE | SCE-UA | 1314 | 2000 | 0.74 | 0.86 |
| C5_seed2 | KGE | SCE-UA | 1329 | 2000 | 0.73 | 0.85 |
| C5_seed0_nse | NSE | SCE-UA | 1339 | 2000 | 0.80 | 0.80 |
| C5_seed1_nse | NSE | SCE-UA | 1302 | 2000 | 0.80 | 0.78 |
| C5_seed2_nse | NSE | SCE-UA | 1327 | 2000 | 0.80 | 0.79 |

**Table B11 Paired VIC search outcomes.** Fixed, DDS after preparation; Agent, continuing agent parameter proposals. Each arm executes 150 calibration evaluations. The difference is validation NSE for fixed minus agent search.

| Block | Fixed calibration NSE | Agent calibration NSE | Fixed validation NSE | Agent validation NSE | Paired difference | Rejected agent proposals |
|---|---|---|---|---|---|---|
| 1 | 0.91 | 0.92 | 0.87 | 0.88 | $-6.89 \times 10^{-3}$ | 1 |
| 2 | 0.91 | 0.92 | 0.87 | 0.88 | $-6.49 \times 10^{-3}$ | 3 |
| 3 | 0.93 | 0.92 | 0.89 | 0.87 | 0.01 | 5 |
| 4 | 0.91 | 0.92 | 0.88 | 0.87 | $1.49 \times 10^{-3}$ | 3 |
| 5 | 0.92 | 0.92 | 0.87 | 0.88 | −0.01 | 1 |
| 6 | 0.91 | 0.92 | 0.87 | 0.87 | $1.35 \times 10^{-3}$ | 5 |
| 7 | 0.92 | 0.91 | 0.87 | 0.87 | $-2.88 \times 10^{-3}$ | 18 |
| 8 | 0.93 | 0.92 | 0.89 | 0.88 | $7.56 \times 10^{-3}$ | 5 |
| 9 | 0.91 | 0.92 | 0.87 | 0.88 | $-8.22 \times 10^{-3}$ | 2 |
| 10 | 0.93 | 0.93 | 0.87 | 0.88 | $-3.11 \times 10^{-3}$ | 18 |

**Table B12 Reproduction of eleven numerical evaluations.** Each package evaluates one specified parameter vector with the archived environment and compares its numerical fields with the original record. Full searches and fresh agent generation were not repeated. Exact agreement uses the saved parameter representation.

| Package | Replayed vector / period | Comparison |
|---|---|---|
| WOFOST | Default seven parameters, five calibration years | Exact numerical fields |
| HBV | Default ten parameters, calibration 2006–2010 | Exact numerical fields |
| SUMMA | Recorded default-like 19-parameter vector, calibration 2006–2010 | Exact numerical fields only with the recorded floating-point vector; clean defaults differ around the ninth significant digit |
| VIC Xixian | Default 15 parameters, calibration 2007–2008 | Exact numerical fields; timing and paths differ |
| MODFLOW 6 | All-one multipliers, calibration FDC | Exact numerical fields |
| CRHM | Default 26 parameters, calibration 2006–2010 | Exact numerical fields; paths and two parameter echoes differ |
| Paired VIC | Last calibration evaluation of fixed-search replicate 1, not its selected best | Exact numerical fields; wall time differs |
| Repeated GR4J | Selected C1_seed0, original KGE cell | Entire metric file byte-identical |
| GR4J reference | Selected best vector, full 1990–1999 | Entire metric file byte-identical |
| SAC-SMA reference | Recorded native parameter-file vector, validation 1989–1999 | All four metrics exactly equal; parameter file byte-identical; the rounded result-file vector gives a tiny mismatch |
| FSM2 | Empty parameters (compiled defaults), validation water years 2019–2020 | Entire metric file byte-identical |

**Table B13 SAC-SMA optimizer means at Ibex.** CAL, calibration; FULL, combined scoring window; VAL, subsequent validation. Each mean describes three numerical seeds under one fixed formulation.

| Method | Nominal budget | Mean actual evaluations | Mean CAL NSE | Mean FULL NSE | Mean VAL NSE |
|---|---|---|---|---|---|
| DDS | 300 | 300.00 | 0.66 | 0.55 | 0.47 |
| SCE-UA | 300 | 390.33 | 0.58 | 0.52 | 0.47 |
| Random | 300 | 300.00 | 0.55 | 0.47 | 0.42 |
| DDS | 1000 | 1000.00 | 0.68 | 0.56 | 0.49 |
| SCE-UA | 1000 | 913.33 | 0.65 | 0.58 | 0.52 |
| Random | 1000 | 1000.00 | 0.61 | 0.51 | 0.45 |
| DDS | 3000 | 3000.00 | 0.69 | 0.57 | 0.50 |
| SCE-UA | 3000 | 1951.67 | 0.68 | 0.59 | 0.53 |
| Random | 3000 | 3000.00 | 0.61 | 0.50 | 0.43 |

**Table B14 SAC-SMA optimizer runs at Ibex.** All 27 runs. The calibration-best parameter vector is selected before validation. Actual evaluations differ from nominal budgets because of optimizer termination behaviour.

| Method | Budget | Seed | Actual evaluations | CAL NSE | FULL NSE | VAL NSE | Wall seconds |
|---|---|---|---|---|---|---|---|
| SCE-UA | 300 | 1 | 388 | 0.60 | 0.57 | 0.55 | 63.60 |
| DDS | 300 | 1 | 300 | 0.66 | 0.54 | 0.47 | 50.90 |
| Random | 300 | 1 | 300 | 0.52 | 0.40 | 0.32 | 50.20 |
| SCE-UA | 300 | 2 | 392 | 0.61 | 0.48 | 0.40 | 66.30 |
| DDS | 300 | 2 | 300 | 0.67 | 0.56 | 0.50 | 51.20 |
| Random | 300 | 2 | 300 | 0.61 | 0.56 | 0.53 | 50.10 |
| SCE-UA | 300 | 3 | 391 | 0.54 | 0.51 | 0.47 | 65.10 |
| DDS | 300 | 3 | 300 | 0.65 | 0.53 | 0.46 | 51.70 |
| Random | 300 | 3 | 300 | 0.51 | 0.45 | 0.41 | 50.90 |
| SCE-UA | 1000 | 1 | 907 | 0.66 | 0.59 | 0.54 | 152.40 |
| DDS | 1000 | 1 | 1000 | 0.67 | 0.57 | 0.51 | 168.30 |
| Random | 1000 | 1 | 1000 | 0.61 | 0.47 | 0.39 | 163.90 |
| SCE-UA | 1000 | 2 | 919 | 0.65 | 0.56 | 0.50 | 151.50 |
| DDS | 1000 | 2 | 1000 | 0.70 | 0.54 | 0.45 | 166.50 |
| Random | 1000 | 2 | 1000 | 0.61 | 0.56 | 0.53 | 166.40 |
| SCE-UA | 1000 | 3 | 914 | 0.64 | 0.58 | 0.53 | 153.20 |
| DDS | 1000 | 3 | 1000 | 0.67 | 0.56 | 0.50 | 166.40 |
| Random | 1000 | 3 | 1000 | 0.60 | 0.50 | 0.44 | 165.80 |
| SCE-UA | 3000 | 1 | 1935 | 0.69 | 0.59 | 0.53 | 319.80 |
| DDS | 3000 | 1 | 3000 | 0.70 | 0.61 | 0.55 | 497.60 |
| Random | 3000 | 1 | 3000 | 0.61 | 0.47 | 0.39 | 499.90 |
| SCE-UA | 3000 | 2 | 1963 | 0.68 | 0.60 | 0.54 | 325.60 |
| DDS | 3000 | 2 | 3000 | 0.69 | 0.51 | 0.40 | 510.10 |
| Random | 3000 | 2 | 3000 | 0.63 | 0.54 | 0.48 | 508.50 |
| SCE-UA | 3000 | 3 | 1957 | 0.69 | 0.58 | 0.52 | 335.60 |
| DDS | 3000 | 3 | 3000 | 0.69 | 0.60 | 0.55 | 509.00 |
| Random | 3000 | 3 | 3000 | 0.60 | 0.50 | 0.44 | 515.80 |

**Table B15 SAC-SMA fitted parameters and reference ranges at the Blanco River.** The generated workflow uses standalone SAC-SMA, Maurer forcing and Hamon potential evapotranspiration; the ten reference seeds of ref. [29] use Daymet forcing and a coupled model. Ranges are the observed seed minima and maxima. The constant reference uzk value of 0.70 does not distinguish fixed from bound-pinned status. lzpk is displayed in $10^{-3}$ $d^{-1}$.

| Parameter | Display unit | Fitted | Reference minimum | Reference maximum | Within reference range? |
|---|---|---|---|---|---|
| uztwm | mm | 34.38 | 11.32 | 11.34 | No |
| uzfwm | mm | 128.21 | 490.04 | 698.21 | No |
| lztwm | mm | 154.62 | 119.46 | 119.93 | No |
| lzfsm | mm | 90.43 | 237.31 | 299.22 | No |
| lzfpm | mm | 131.55 | 457.38 | 577.32 | No |
| uzk | $d^{-1}$ | 0.84 | 0.70 | 0.70 | No |
| lzpk | $10^{-3}$ $d^{-1}$ | 13.24 | 10.00 | 11.00 | No |
| lzsk | $d^{-1}$ | 0.26 | 0.15 | 0.15 | No |
| zperc | dimensionless | 207.99 | 80.78 | 118.39 | No |
| rexp | dimensionless | 3.24 | 3.86 | 4.97 | No |
| pfree | fraction | 0.33 | 0.10 | 0.11 | No |

**Table B16 Language-model roles in related scientific workflows.** The table describes the reported architectures and evaluations of the studies cited in the main text: Zhu et al. [12], Yan et al. [13], Li et al. [14], Yang et al. [15], Zhang et al. [19] and Li et al. [18].

| Study | Where the language model acts | Model or task scope | Knowledge and tools supplied | Evaluation summarized here | Evaluation scope |
|---|---|---|---|---|---|
| Zhu et al. [12] | Inside iterative calibration, proposing parameter updates from simulation diagnostics | VIC; five tested language models | Parameter meanings and ranges, role and objective prompts, diagnostic metrics and hydrographs | Calibration performance and convergence compared with SCE-UA and NSGA-III | Parameter meanings, ranges and diagnostics ground the tested parameter proposals; performance varies across the five language models |
| Yan et al. [13] | Workflow orchestration: request interpretation, data retrieval, model configuration, simulation and reporting, with human oversight | CREST-based flood modelling; Kerr County repeated workflow and selected other test events | Model and data tools, event interpretation, structured workflow and user confirmation | First-pass workflow completion in 18 of 20 Kerr County trials; hydrologic accuracy assessed separately with streamflow metrics | The 18/20 result measures workflow completion for one repeated case; hydrologic accuracy is evaluated separately, and other selected events extend the application examples |
| Li et al. [14] | As a trained policy proposing calibration parameter sets | CREST through the EF5 simulator; frontier-agent benchmark and a post-trained open model | Supervised calibration trajectories followed by reinforcement learning with simulator feedback; explicit parameter and diagnostic tools | Calibration benchmark and simulator-feedback policy training | Evaluation is specific to the simulator, calibration tasks and post-training protocol; the agent acts as a parameter-proposal policy |
| Yang et al. [15] | Agent reasoning within explicit forecasting skills; numerical tools perform calibration and simulation; selected steps receive human review | Xinanjiang flood forecasting | Codified forecaster expertise, skill rules, structured JSON artefacts and a DDS calibration tool | Skill-orchestrated flood forecasting workflows | Explicit forecasting skills separate agent judgement from numerical tools and retain inspectable procedures |
| Zhang et al. [19] | Assistive reasoning, programming and forecasting tasks | Hydrometeorological tasks with GPT-4o, Claude 3.5 Sonnet and Gemini 1.5 Pro | Task prompts, data and programming tasks | Data-processing successes alongside incorrect method selection on some scientific questions and limited forecasting performance | The tested tasks include scientific method-selection errors; the study is an assistive hydrometeorological assessment rather than a calibration-specific benchmark |
| Li et al. [18] | Agentic scientific simulation guided by reusable operational knowledge | Multiple Earth-science simulation tasks | Validated modelling operators, staged domain protocols and diagnostic recovery mechanisms | Cross-model applications and re-executability of scientific workflows | Operational knowledge supports simulation tasks, reusable procedures and numerical re-execution across model applications |
| This study | Formulation of calibration specifications and runners before numerical search; a separate paired comparator proposes parameters inside the search | Six process-based model families, with focused GR4J, FSM2, VIC, HYMOD and SAC-SMA tests | Knowledge infrastructure with interfaces, parameter applicability and ranges, observation mappings, assessment conventions, tools and diagnostics; implemented checks and review | Agent-led six-model preparation; supplied-knowledge tests of first-pass usability and skill; repeated GR4J formulations; paired VIC searches; reference comparisons and numerical replays | Agent-led adaptation across model families is complemented by focused preparation, reference, consistency and replay tests; generation uses one language-model configuration and a second configuration serves the continuing-proposal comparison |

**Table B17 Direct-driver multivariable applications.** Four hand-written driver applications assess numerical execution independently of agent generation. Component losses are given in Table B18.

| Case | Parameters / objectives | Requested budget; seed | Saved evaluations / front size | Accepted objectives at selected point |
|---|---|---|---|---|
| WOFOST KBS yield + annual soil moisture | 4 / 3 | 40; 0 | 40 / 4 | 1/3 |
| WOFOST KBS yield + GlobMap LAI | 5 / 3 | 60; 0 | 80 / 13 | 0/3 |
| CRHM Dore discharge + CanSWE | 26 / 4 | 120; 0 | 120 / 19 | 3/4 |
| VIC Tangnaihai discharge + MOD16 ET | 15 / 4 | 24; 0 | 40 / 23 | 2/4; no finite ET validation |

**Table B18 Component losses in the direct-driver applications.** Loss definitions are specific to each component. Joint component acceptance follows the configured scientific assessment; unavailable values indicate absent finite validation metrics. SM, soil moisture; LAI, leaf-area index.

| Case / objective | Calibration loss | Validation loss | Baseline validation loss | Joint acceptance |
|---|---|---|---|---|
| WOFOST SM: yield magnitude | 0.10 | 0.32 | 0.19 | No |
| WOFOST SM: SM temporal | 3.11 | 0.77 | 8.93 | No |
| WOFOST SM: SM magnitude | 0.02 | 0.03 | 0.26 | Yes |
| WOFOST LAI: yield magnitude | 0.23 | 0.32 | 0.33 | No |
| WOFOST LAI: LAI temporal | 1.74 | 1.64 | 1.42 | No |
| WOFOST LAI: LAI magnitude | 0.08 | 0.22 | 0.17 | No |
| CRHM: Q magnitude | 0.02 | 0.05 | 0.07 | Yes |
| CRHM: Q temporal | 0.76 | 0.77 | 0.29 | No |
| CRHM: SWE magnitude | 0.06 | 0.05 | 0.10 | Yes |
| CRHM: SWE temporal | 0.07 | 0.05 | 0.09 | Yes |
| VIC: Q temporal | 0.33 | 0.24 | 0.35 | Yes |
| VIC: Q magnitude | 0.01 | 0.03 | 0.12 | Yes |
| VIC: ET temporal | 3.52 | Not available | Not available | No |
| VIC: ET magnitude | 0.06 | Not available | Not available | No |

**Table B19 HYMOD optimizer endpoints at Leaf River.** Full-window NSE at six nominal budgets and three seeds. Reported repetitions are optimizer counters and can exceed nominal budgets; they do not enumerate all physical model calls. Source data retain full available precision.

| Algorithm | Nominal budget | Seed or aggregation | Reported repetitions | Best NSE |
|---|---|---|---|---|
| DDS | 200 | 0 | 200 | 0.68 |
| DDS | 200 | 1 | 200 | 0.68 |
| DDS | 200 | 2 | 200 | 0.68 |
| SCE-UA | 200 | 0 | 200 | 0.64 |
| SCE-UA | 200 | 1 | 200 | 0.57 |
| SCE-UA | 200 | 2 | 200 | 0.59 |
| DDS | 500 | 0 | 500 | 0.68 |
| DDS | 500 | 1 | 500 | 0.68 |

| Algorithm | Nominal budget | Seed or aggregation | Reported repetitions | Best NSE |
|---|---|---|---|---|
| DDS | 500 | 2 | 500 | 0.68 |
| SCE-UA | 500 | 0 | 756 | 0.64 |
| SCE-UA | 500 | 1 | 743 | 0.61 |
| SCE-UA | 500 | 2 | 757 | 0.60 |
| DDS | 2000 | 0 | 2000 | 0.68 |
| DDS | 2000 | 1 | 2000 | 0.68 |
| DDS | 2000 | 2 | 2000 | 0.68 |
| SCE-UA | 2000 | 0 | 2503 | 0.67 |
| SCE-UA | 2000 | 1 | 2467 | 0.67 |
| SCE-UA | 2000 | 2 | 2511 | 0.68 |
| DDS | 5000 | 0 | 5000 | 0.68 |
| DDS | 5000 | 1 | 5000 | 0.68 |
| DDS | 5000 | 2 | 5000 | 0.68 |
| SCE-UA | 5000 | 0 | 5472 | 0.68 |
| SCE-UA | 5000 | 1 | 5468 | 0.68 |
| SCE-UA | 5000 | 2 | 5502 | 0.68 |
| DDS | 10000 | 0 | 10000 | 0.68 |
| DDS | 10000 | 1 | 10000 | 0.68 |
| DDS | 10000 | 2 | 10000 | 0.68 |
| SCE-UA | 10000 | 0 | 10242 | 0.68 |
| SCE-UA | 10000 | 1 | 10227 | 0.68 |
| SCE-UA | 10000 | 2 | 10302 | 0.68 |
| DDS | 20000 | 0 | 20000 | 0.68 |
| DDS | 20000 | 1 | 20000 | 0.68 |
| DDS | 20000 | 2 | 20000 | 0.68 |
| SCE-UA | 20000 | 0 | 20483 | 0.68 |
| SCE-UA | 20000 | 1 | 20447 | 0.68 |
| SCE-UA | 20000 | 2 | 20550 | 0.68 |

**Table B20 HYMOD DDS comparator for the random-search comparison.** The nine DDS entries are the same seeded searches represented in Table B19 and provide the comparator for Table B21.

| Algorithm | Nominal budget | Seed or aggregation | Reported repetitions | Best NSE |
|---|---|---|---|---|
| DDS | 500 | 0 | 500 | 0.68 |
| DDS | 500 | 1 | 500 | 0.68 |
| DDS | 500 | 2 | 500 | 0.68 |
| DDS | 2000 | 0 | 2000 | 0.68 |
| DDS | 2000 | 1 | 2000 | 0.68 |
| DDS | 2000 | 2 | 2000 | 0.68 |
| DDS | 5000 | 0 | 5000 | 0.68 |
| DDS | 5000 | 1 | 5000 | 0.68 |
| DDS | 5000 | 2 | 5000 | 0.68 |

**Table B21 HYMOD random-search means.** Each value is the mean over seeds 0, 1 and 2. Individual random-search endpoints and uncertainty estimates are unavailable. Scoring uses the full window.

| Algorithm | Nominal budget | Seed or aggregation | Reported repetitions | Best NSE |
|---|---|---|---|---|
| Random | 500 | Mean of seeds 0, 1, 2 | Unavailable | 0.62 |
| Random | 2000 | Mean of seeds 0, 1, 2 | Unavailable | 0.65 |
| Random | 5000 | Mean of seeds 0, 1, 2 | Unavailable | 0.66 |

**Table B22 Parameter-range diagnostics.** One numerical seed per arm in five model families. A failed sensitivity screen yields no assessable calibrated outcome; inconclusive split validation is reported separately from scientific rejection. These diagnostics use their original assessment rules rather than the common six-model convention.

| Model | Supplied knowledge-infrastructure ranges | Widened ranges | Observed effect |
|---|---|---|---|
| WOFOST | Sensitivity screen returned no finite losses; no assessable calibrated outcome | 80 recorded optimizer evaluations; grain NRMSE 28.73%; biomass NRMSE 29.57%; 1/2 accepted; both NRMSE values ≤ 30.00% | Widening permits an assessable result after the supplied-range screen fails; one component is accepted |
| HBV | 160 recorded optimizer evaluations; Q NSE 0.50, absolute bias 25.77%; SWE NSE 0.86, absolute bias 3.41%; absolute mean SWE peak-date bias 8.60 days; 3/5 accepted | 160 recorded optimizer evaluations; Q NSE 0.03, absolute bias 25.18%; SWE NSE 0.83, absolute bias 4.68%; absolute mean SWE peak-date bias 8.60 days; 2/5 accepted | Supplied ranges preserve higher Q and SWE NSE; Q bias remains above 15.00% in both arms |
| SUMMA | 160 recorded optimizer evaluations; Q NSE 0.69, absolute bias 1.02%; SWE NSE 0.79, absolute bias 15.05%; 4/4 accepted | 160 recorded optimizer evaluations; Q NSE 0.74, absolute bias 1.94%; SWE NSE 0.82, absolute bias 5.68%; 4/4 accepted | Widening improves three of four validation losses; all four components pass in both arms |
| MODFLOW 6 | 160 recorded optimizer evaluations; Q FDC NSE 0.86, absolute bias 15.50%; baseflow FDC NSE 0.69, absolute bias 34.88%; 2/4 accepted | 160 recorded optimizer evaluations; calibration completed, but the split probe left validation inconclusive (metric response within 2.00% on both splits) | The widened search completes, but the split probe leaves validation inconclusive; the same simulated FDC is compared with two observed FDCs |
| CRHM | 160 recorded optimizer evaluations; Q NSE −0.24, absolute bias 18.09%; SWE NSE 0.68, absolute bias 10.89%; 3/4 accepted | 160 recorded optimizer evaluations; Q NSE −0.14, absolute bias 7.67%; SWE NSE −0.59, absolute bias 83.24%; 1/4 accepted | Supplied ranges preserve SWE fit; widening improves Q losses while Q temporal acceptance remains unmet |

**Table B23 Calibration-only cached selection diagnostics.** Pairs are NSE / absolute PBIAS (%). The alternative selects the existing candidate minimizing mean max(0, 1 − NSE) across targets. Both columns use calibration records. Crop NSE and bias are diagnostic metrics; the original optimization used NRMSE.

| Model and variables | Minimum mean clipped NSE loss, calibration rows only | Report-selected member's calibration metrics |
|---|---|---|
| WOFOST TWSO / TAGP | 0.73 / 14.63%; 0.46 / 1.02% | 0.85 / 6.56%; 0.18 / 10.76% |
| HBV Q / SWE | 0.63 / 20.16%; 0.90 / 9.12% | 0.60 / 3.54%; 0.89 / 16.72% |
| SUMMA Q / SWE | 0.75 / 9.03%; 0.89 / 5.77% | 0.58 / 3.46%; 0.88 / 1.17% |
| VIC Q / ET | 0.59 / 66.53%; 0.89 / 1.42% | 0.85 / 0.40%; −0.32 / 36.44% |
| MODFLOW 6 Q / baseflow FDC | 0.92 / 0.44%; 0.82 / 6.04% | 0.88 / 7.62%; 0.78 / 1.29% |
| CRHM Q / SWE | 0.48 / 14.23%; 0.36 / 49.10% | 0.46 / 0.57%; 0.25 / 53.46% |

**Table B24 Numerical-search comparison definitions and outcomes.** Hypervolumes (HV) use experiment-specific reference points. Differences may remain visible at source precision when the two-decimal values coincide. Fig. A6 shows the signed within-experiment differences.

| Experiment | Objectives and reference | Nominal effort | Result | Comparison scope |
|---|---|---|---|---|
| HYMOD weighted-scalar comparison | [absolute PBIAS / 100, 1 − r]; full scoring window; HV reference (1.00, 1.50) | NSGA-II: T; five DDS searches: T / 5 each, weights 0.15, 0.35, 0.50, 0.65, 0.85; T = 100, 300, 600 | NSGA-II versus DDS pooled candidates: T = 100, HV 1.23 versus 1.28 (difference $-4.80 \times 10^{-2}$); T = 300, 1.32 versus 1.32 ($+1.00 \times 10^{-3}$); T = 600, 1.32 versus 1.32 ($+1.00 \times 10^{-3}$) | Retained HV is printed to three decimals; the larger-budget contrasts are near ties at that precision; no independent validation or replicated superiority inference |
| WOFOST unequal nominal budgets | Validation projection [absolute yield PBIAS / 100, 0.50 max(0, 1 − soil-moisture NSE) + 0.50 absolute soil-moisture PBIAS / 100]; data-dependent HV reference (1.20, 1.95), rounded here | NSGA-II: 40; five DDS searches: 30 each, total 150; weights 0.10, 0.30, 0.50, 0.70, 0.90 | HV 1.39 versus 1.55; exact retained reference and HV values are in the source record | Unequal nominal effort; the reference point differs from the equal-budget WOFOST comparison |
| WOFOST equal nominal budgets | Same two-coordinate validation projection; HV reference (1.20, 2.20); NSGA-II searches three objective families | NSGA-II: T; five DDS searches: T / 5 each, weights 0.15, 0.35, 0.50, 0.65, 0.85; T = 60, 180, 360 | NSGA-II versus DDS pooled candidates: T = 60, HV 1.68 versus 1.33 (difference +0.35); T = 180, 1.75 versus 1.76 ($-7.00 \times 10^{-3}$); T = 360, 1.80 versus 1.62 (+0.18) | Candidates arise from calibration searches and are projected onto validation metrics; retained HV is printed to three decimals; ordering changes with budget; exact realized model-call parity is unverified |
| WOFOST three-backend comparison | Three recorded objective-family assessments: yield magnitude, soil-moisture temporal fit and soil-moisture magnitude; no HV comparison | NSGA-II, NSGA-III and MOEA/D: nominal budget 40 each | Retained Pareto members: 4, 3, 23; accepted components: 1/3, 2/3, 2/3; no run satisfies all component criteria | Actual total model-call counts are unavailable in the retained summary record |
| CRHM knee diagnostic | One engine-reported validation point [discharge temporal loss, SWE magnitude loss], compared with the previous scalar and random candidates; no full-front HV | New NSGA-II executions at nominal T = 60 and 180; previous scalar and random candidate sets reused | T = 60: knee (0.39, 0.23), dominated by a scalar candidate; T = 180: knee (0.28, 0.09), not dominated by a scalar candidate; neither knee is dominated by the retained random candidates | Single-point diagnostic; random candidates were scored directly on validation, so this is not a matched predictive-performance comparison |

## Appendix C: The harness, its checks and its numerical settings

### Implementation checks

Before any search, the harness certifies the generated runner against the validated reference run of the knowledge infrastructure (Methods). During search it checks candidate constraints before execution and compares each requested parameter vector with the runner's applied-parameter report, and the runner reads the applied values back from the native files the model consumes. Case-identity checks apply when a case is pinned, and the individual runners check native files, output length, gauge identity and input hashes. Eighteen commissioning, parameter-response and split-probe records cover five model families: parameter perturbations change the simulated outputs, and split-specific requests produce distinct scored observation records.

With the model graph and the validation convention a knowledge infrastructure becomes a verifiable environment in the sense used for agent benchmarks such as SWE-bench [16]: the agent formulates from the same description against which its formulation is checked. In the FSM2 experiment this pre-search check rejected every name-only specification and two raw-documentation specifications before any model run.

### Calibration and validation data use

The six-family application selects a final front member by exhaustive assessment of validation losses, with discharge protection for HBV, VIC and CRHM, so its validation data inform solution selection. FSM2 and the paired VIC experiment select parameters on calibration data alone and then evaluate validation performance. The repeated-formulation experiment and the GR4J reference comparison score the full optimization window without a separate validation period, as do the HYMOD numerical comparisons.

### Search budgets

Budget allocation is part of formulation. In the authoring tests, agents declared nominal budgets from the guidance in the knowledge infrastructure, the number of active parameters and the expected model cost, supplemented by their own heuristics. FSM2 specification K1, for example, proposed 300 DDS proposals for five parameters from a one-second run-cost estimate and a heuristic of about 60 proposals per parameter, which the agent labelled as its own prior knowledge; the repeated GR4J formulations referred to the typical range of 1,000–5,000 evaluations recorded in the knowledge infrastructure and selected 1,500–3,000. The paired VIC comparison prescribed 150 calibration evaluations per arm and the optimizer comparisons used fixed budget grids.

The current harness also budgets from a wall-clock allowance. The agent declares either a fixed cap with a rationale or an available duration. For the latter, the harness measures evaluation cost with a ten-run pilot, or a three-run pilot when the user supplies a runtime estimate, and derives a per-seed cap from a conservative runtime estimate, overhead and available parallelism; the conservative estimate is 1.2 times the 90th-percentile pilot runtime, or 1.2 times the larger of the supplied estimate and the slowest pilot run, and three seeds are used by default.

### Convergence

Convergence is assessed separately from the budget and from scientific acceptance. At completed SCE-UA evolution loops the harness checks a normalized geometric population range below 0.001, or a change of at most 0.1% over the last ten loops in the best objective and in the watched scores (correlation, alpha, beta and logarithmic NSE). For NSGA-II, NSGA-III and MOEA/D it applies the design-space and objective-space termination tests of pymoo with xtol = 0.0005, ftol = 0.005, period = 50 and n_skip = 5, excluding the evaluation and generation caps from the decision. DREAM retains its native criterion of R-hat below 1.2. DDS has no native convergence test, so the harness reports the last material change within the search as a descriptive settle point and preserves the budget-dependent DDS schedule. The default mode records diagnostics while the search continues; an explicit stop mode terminates SCE-UA or pymoo searches at a completed loop or generation. Records distinguish the reason execution ended from the converged, not converged or unknown verdict, and across-seed convergence additionally requires agreement in the watched final calibration scores. Requested caps, optimizer counters and logged evaluations, including cache hits and rejected candidates, are recorded separately and need not equal completed model executions. This budgeting and convergence implementation is in GCH commit 5722f2b (4 October 2026; https://github.com/lzwei196/GCH); the reported experiments used the archived engine and search settings.

## Appendix D: Adaptation across six model families

### Models, sites and observations

The six families cover crop growth (WOFOST), lumped rainfall–runoff (HBV), land-surface (SUMMA), macroscale hydrology (VIC), groundwater (MODFLOW 6) and cold-region hydrology (CRHM), with different native interfaces. Sites were chosen where two observed variables are available. WOFOST was run at Kellogg Biological Station, where in-situ grain mass and destructive biomass are both measured; gridded leaf-area and soil-moisture products were tested earlier and were too coarse (Tables B17 and B18). HBV and SUMMA were run at Moyie, a snow-dominated catchment with a CanSWE pillow, so that both models see the same discharge and snow records. VIC was run at Xixian, where monthly GLEAM evapotranspiration is available; Tangnaihai, the reference case of the VIC knowledge infrastructure, was reserved for the paired-search experiment. MODFLOW 6 used the Sagehen example distributed with the model, which has a co-located gauge but no head observations, so streamflow and filtered baseflow were the targets. CRHM was first run at Dore, where the uncalibrated model already met the criteria, and then at Blue River, which has a gauge, a snow pillow and a poor uncalibrated baseline. Model configurations, periods and parameter counts are given in the Methods.

## Agent-led formulation and numerical execution

A coordinating agent and model-specific agents prepared the six formulations and commissioned them at budgets of 40–200 evaluations under human direction of the campaign's scope and changes of case. Numerical execution then used the budgets declared in the contracts and the common assessment convention. All six declared-budget searches used NSGA-II with seed 0. The HBV and MODFLOW 6 specifications included an optimizer rationale; SUMMA and VIC requested DDS, which the harness replaced with NSGA-II for the multiobjective task. Nominal budgets were 600 (WOFOST), 500 (HBV), 240 (SUMMA), 150 (VIC), 200 (MODFLOW 6) and 500 (CRHM); population termination produced 520 optimizer evaluations for HBV and CRHM. WOFOST activated three, five and seven parameters in three stages and VIC increased its active set from three to thirteen parameters across six stages. The saved results contain 2,536 distinct calibration vectors, 385 validation records and 167 final-front members, of which validation records cover 157; MODFLOW 6 selection stopped when the second assessed member passed all components. Sixteen of 23 components satisfied joint acceptance, and one of the six workflows, MODFLOW 6, satisfied every component (Fig. 4; Tables B8 and B9). Five flow cases met their selected numerical flow thresholds; MODFLOW 6 assessed duration curves and the other four assessed daily discharge. Commissioning results are listed in Table B7. WOFOST commissioning used absolute PBIAS and a maximum-biomass operator, whereas the declared-budget experiment used NRMSE and final-maturity biomass, so the commissioning and declared-budget results are distinct formulations.

## Workflow-generation examples

Two further generated workflows illustrate the scientific content of the agents' choices. A WOFOST example at Kellogg Biological Station produced a specification with nineteen candidate parameters and applicability notes for water-limited production; its PCSE runner converted dry grain yield to the 15% moisture basis of the observations and gravimetric soil moisture to volumetric units with a bulk density of 1.41. Morris screening and a staged multiobjective design activated six parameters (SMFCF, SMW, WAV, CFET, TSUM1 and SOW_OFFSET_DAYS) at a requested budget of forty evaluations; both soil-moisture components passed the configured assessment while yield failed its magnitude criterion. A CRHM example generated eighteen parameters and a native runner with an NSGA-II rationale and completed preparation after one revision; objective probing removed two timing families absent from the model output, and because the baseline discharge NSE of 0.70 met the configured adequacy criterion the workflow stopped before optimization. A separate decision probe supplied two observed variables combined into one scalar score, and the agent specified a single objective and selected DDS.

## Direct-driver multivariable applications

Four hand-written drivers exercised the multivariable support of the harness independently of agent generation (Tables B17 and B18). WOFOST yield with annual soil moisture used four parameters, three objectives, a budget of forty and seed 0, with one of three components accepted. WOFOST yield with GlobMap leaf-area index used five parameters and a nominal budget of sixty, produced eighty evaluations and accepted no component. CRHM Dore discharge with CanSWE snow used 26 parameters and a budget of 120 and accepted three of four components, with snow NSE reaching 0.95 while discharge NSE fell from 0.71 to 0.23. VIC Tangnaihai discharge with MOD16 evapotranspiration used fifteen parameters and a nominal budget of 24, produced forty evaluations and raised discharge NSE from 0.65 to 0.76, with no validation evapotranspiration because the observations did not cover that period. All four completed and none satisfied every component criterion.

### Single-objective phase, authoring cost and cross-vendor authoring

Before the multivariable campaign, each family was calibrated once as a single-objective task (Table B3). An agent read the knowledge infrastructure and authored a specification and runner for one observed variable; each specification was reviewed by an independent model before execution and was executed unchanged at 80 evaluations. Five specifications selected DDS and CRHM selected SCE-UA. WOFOST, MODFLOW 6 and VIC completed calibration and met the holdout criterion; HBV, SUMMA and CRHM stopped before optimization because the default parameters already met the declared target. Applied to basins with poor default fit, DDS at 120 evaluations raised holdout NSE from 0.59 to 0.73 for HBV and from 0.40 to 0.61 and 0.70 for two SUMMA seeds, and at Belly River three DDS seeds at 600 evaluations raised CRHM holdout NSE from 0.44 to 0.71–0.73 while three SCE-UA seeds reached 0.67–0.71. Model evaluation time spanned 6.9 s (WOFOST) to 1,074 s (VIC) per run.

One authoring by claude-fable-5 used 16,575–87,042 output tokens, cost US$3.57–14.21 and took 11–34 min (mean 53,262 tokens, US$10.05 and 23 min; Table B4). The WOFOST, HBV and MODFLOW 6 tasks were re-authored by claude-sonnet-5, claude-opus-4-8 and GPT-5.5; all nine specifications were runnable and selected DDS, so fifteen of fifteen authorings across four configurations from two vendors produced runnable single-objective workflows, with Sonnet and Opus authorings costing US$2.65–4.47. Kimi and DeepSeek lacked headless authoring support in the available tooling and were not tested.

Two decision tests examined whether the optimizer choice follows the objective structure. In knowledge-grounded authoring, HBV, VIC and MODFLOW 6 each selected DDS when the deliverable was a point estimate and DREAM when it was parameter uncertainty; SUMMA selected DDS and CRHM selected SCE-UA under the point-estimate framing. In a single-shot blinded test offering DDS, SCE-UA, DREAM and NSGA-II, Claude, GPT and Kimi agreed on twelve of twelve choices across four framings (Table B5); all three chose DDS rather than the textbook SCE-UA for a rugged surface at moderate budget, which the Belly River comparison supports. The CRHM authoring that selected SCE-UA is the one recorded within-class mismatch.

The working directory of these single-objective runs was cleared by a server reboot on 24 August 2026. Their specifications are retained with their hashes, and their numerical outputs survive as the console output captured in the coordinator session record, from which Tables B3–B5 are transcribed; the runs therefore appear in no figure.

## Appendix E: Knowledge experiments and reference comparisons

### FSM2 information conditions and calibration protocol

The preregistered knowledge experiment used FSM2 [24] at Mt Rose (SNOTEL 652). The structured-knowledge condition supplied twenty files (226,250 bytes), the raw-documentation condition nine files (92,822 bytes) and the name-only condition the model name and the common task schema. Five independent generation sessions per condition completed on their first attempt, and because a fixed runner was supplied the agent generated only the specification; the K and R packages differed in content and volume as well as in organization. DDS used seed 0 and each specification's budget of 200–500 proposals over a calibration window of water years 2016–2018 (1,096 days) and a validation window of water years 2019–2020 (731 days). The compiled defaults gave validation NSE 0.57 and KGE 0.43. The predeclared primary outcomes were execution without revision and validation NSE above the defaults, and the support criterion required at least four of five K and at most one of five M specifications to execute without revision. Execution without revision succeeded for five of five K, three of five R and none of five M specifications; K calibrations reached validation NSE 0.81–0.94; and a two-sided Fisher exact comparison of first-pass execution between K and M gave $P = 7.90 \times 10^{-3}$. Improvement is measured against the compiled defaults.

### Prespecified repairs and generation effort

The repair protocol first corrected the target name from SWE to the exposed variable snw and then converted the hourly albedo and compaction timescales tmlt, tcld and trho to seconds by multiplication by 3,600. Two M specifications also contained bstb, a parameter outside the supplied interface and the repair scope, and received no repair search. Repairs reused the initial generations. After correction, all five R specifications reached validation NSE 0.96–0.98 and the three eligible M specifications 0.93–0.99, and the unit correction improved validation NSE in all eight eligible pairs. The experiment comprised fifteen generations and 21 searches, eight initial and thirteen repair. Of 6,350 proposals, 6,348 were scored; two R2 proposals violated constraints; caching and constraints reduced completed model executions to 5,653 (Table B6). The fifteen generation sessions cost US$12.66 in total, with a median elapsed session time of 3.00 min and a summed session time of 43 min under parallel execution; these values describe generation only. An exploratory precursor of nine generations, three per condition, informed the confirmatory design: initial execution succeeded for three of three K, one of three R and none of three M specifications, and three target-corrected searches exceeded their own weaker defaults but not the common compiled-default reference. Exploratory and confirmatory observations were analysed separately.

### GR4J parameter ranges and observation mapping

Nine fresh GR4J specifications, three per condition, were assessed without calibration. The K package contained sixteen files (155,005 bytes), the R package twenty files (230,674 bytes) and the M condition the model name, the task and the common schema. Each proposed transformed range was compared by intersection over union with the empirical 10th–90th percentile box of the developers [25]: X1 [100, 1,200], X2 [−5, 3], X3 [20, 300] and X4 [1.10, 2.90] (Fig. A2). Overlap was 0.81–1.00 for K, 0.41–0.53 for R and 0.39–0.60 for M. All nine specifications recovered X1–X4 and included the reference optimum. All K and R specifications selected the exposed runner target Qsim, whereas all M specifications selected Q, which the runner did not expose; M specifications also used a logarithmic X4 transform where airGR uses an affine convention. The evaluated outcome is conformity to the supplied interface and to the empirical ranges.

### GR4J reference calibration

The generated workflow was evaluated at basin L0123001 (360 km$^2$) with 1989 as warm-up and 1990–1999 as the scoring window, comprising 3,595 valid observations among 3,652 dates. DDS with seed 0 used 1,000 evaluations and the airGR Calibration_Michel reference [28] used 234; both achieved NSE 0.80 and KGE 0.79. The generated parameters were 253.65, 0.97, 87.55 and 2.21 against reference values of 257.24, 1.01, 88.23 and 2.21 (Fig. A5a,b). The reference hydrograph was unavailable, so the comparison is of scores and parameters.

### SAC-SMA fit and parameter references

A knowledge-guided agent generated a standalone SAC-SMA workflow for the Blanco River (USGS 08171300) with Maurer forcing and Hamon potential evapotranspiration, eleven free and five fixed parameters. The authoring instructions supplied the case, the periods and the reference-method information, so agreement with published choices is descriptive. SCE-UA with seed 0 used 2,702 evaluations from a declared nominal budget of 3,000. Calibration from 1 October 1999 to 30 September 2008 gave NSE 0.72, KGE 0.66 and PBIAS +26.17%; validation from 1 October 1989 to 30 September 1999 gave NSE 0.70, KGE 0.69 and PBIAS +24.82%. The published CAMELS benchmark [29] gave validation NSE 0.63, KGE 0.70 and PBIAS −0.53% with a coupled Snow-17/SAC-SMA/unit-hydrograph model and a different potential-evapotranspiration scheme (Fig. A5c). Parameter choices were compared with ten Daymet-forced reference seed files of ref. [29] for the same basin: ten empirically varying parameter identities matched; uzk was constant at 0.70 across the files; none of the eleven fitted values lay inside its reference range; four of five fixed values matched; and the supplied uzfwm bounds of 5.00–150.00 mm excluded the reference values of 490.04–698.21 mm (Fig. A5d; Table B15). The reference files use Daymet forcing and a coupled model, whereas the generated workflow uses Maurer forcing and standalone SAC-SMA.

## Appendix F: Consistency and numerical reproducibility

### Repeated GR4J formulation

Five fresh knowledge-guided GR4J specifications were each executed with optimizer seeds 0, 1 and 2. Four selected SCE-UA and one selected DDS, with nominal budgets of 1,500–3,000 evaluations; SCE-UA terminated after 1,302–2,135 evaluations. Specifications C1 and C5 selected KGE, and each of their seeded searches was also executed with NSE substituted as the objective, producing six matched interventions and 21 searches in total (Table B10). All searches used a common runner and scored the full 1990–1999 window of 3,595 valid days without a separate validation period. The fifteen NSE endpoints, nine from the original NSE specifications and six from the interventions, had a total NSE range of $9.22 \times 10^{-4}$ and all rounded to 0.80; the largest within-specification seed range was $2.29 \times 10^{-4}$; replacing KGE with NSE shifted mean NSE by about 0.06.

### Fixed search and continuing agent proposals

The preregistered VIC experiment at Tangnaihai fixed a six-parameter formulation with loss 1 − NSE, calibration over 2007–2011 and validation over 2012–2016. Ten paired blocks compared DDS with an agent proposing successive parameter vectors from simulation diagnostics. Each block shared five starting points, the runner default and four Latin-hypercube points; DDS generated 145 further proposals without language-model calls, and the agent arm supplied 145 accepted proposals with 1–18 rejected proposals per block. Both arms executed exactly 150 calibration evaluations and one validation evaluation of the calibration-selected vector. The estimand was the median paired difference in validation NSE, fixed minus continuing, with a 95% interval from 10,000 complete-block bootstrap resamples (seed 0) and a prespecified non-inferiority margin of −0.02. Median validation NSE was 0.87 for fixed search and 0.88 for continuing proposals; the paired median difference was $-2.99 \times 10^{-3}$ with interval [$-7.35 \times 10^{-3}$, $2.34 \times 10^{-3}$], which satisfies the margin (Table B11). Ten continuing-agent attempts interrupted by transport errors, rate limits or a host reboot were excluded under the preregistered rules; no fixed-search attempt failed.

### Reproduction of specified numerical evaluations

Eleven replay packages each evaluated one specified parameter vector and compared the numerical outputs with the saved result (Table B12). They cover the six family applications, the paired VIC experiment, the repeated GR4J formulation, the two reference comparisons and FSM2. All reproduced their comparison metrics, and the repeated-GR4J, GR4J-reference and FSM2 metric files were byte-identical. Several packages evaluate default vectors; the recorded floating-point vector was required for SUMMA and the native parameter-file vector for the SAC-SMA reference to retain exact agreement.

## Appendix G: Numerical strategies and scientific diagnostics

### SAC-SMA optimizer comparison

The optimizer comparison used a fixed literature-derived SAC-SMA setup at Ibex (HYDAT 09AC007) and a dedicated SPOTPY driver, without knowledge-guided authoring: five equal-area hydrologic response units, degree-day snow and Hamon potential evapotranspiration, simulation over 2005–2015 with calibration over 2006–2010 and validation over 2011–2015, and thirteen free parameters. DDS, SCE-UA [27] and uniform random search were compared at nominal budgets of 300, 1,000 and 3,000 with seeds 1, 2 and 3, giving 27 fixed-formulation searches and 35,566 finite saved evaluations; the calibration-best vector was selected before validation scoring. SCE-UA realized 388–392, 907–919 and 1,935–1,963 evaluations at the respective nominal budgets. At a nominal budget of 3,000, mean validation NSE was 0.53 for SCE-UA, 0.50 for DDS and 0.43 for random search, at mean actual effort of about 1,952, 3,000 and 3,000 evaluations (Tables B13 and B14; Fig. A4).

### HYMOD and multivariable search comparisons

HYMOD at Leaf River was evaluated with DDS and SCE-UA at nominal budgets of 200, 500, 2,000, 5,000, 10,000 and 20,000, each with seeds 0–2, producing 36 endpoints scored on the full available window (Fig. A6a; Table B19). SCE-UA repetition counters sometimes exceeded the nominal budget and do not enumerate every model call. The random-search comparison uses one three-seed mean at each of three budgets, and its nine DDS entries are seeded searches already in the main sweep (Fig. A6b; Tables B20 and B21).

Weighted-scalar and multiobjective searches were compared for HYMOD and WOFOST with seed 0, each scalar pool combining five DDS searches at one fifth of the NSGA-II nominal budget. HYMOD used the full-window objectives [absolute PBIAS/100, 1 − correlation]; WOFOST optimized three objective families and was assessed through a two-axis validation projection; hypervolume references and objectives are defined in Table B24 and the signed differences are shown in Fig. A6c,d. WOFOST was also run with NSGA-II, NSGA-III and MOEA/D at a nominal budget of forty and seed 0; final-front sizes were four, three and 23 and accepted components one, two and two of three, so no backend satisfied all component criteria (Fig. A6e,f). A separate CRHM diagnostic compared one engine-reported validation knee at each of two budgets with existing scalar and random candidates: the knee was dominated by a scalar candidate at a budget of 60 and was not dominated at 180, and neither knee was dominated by the retained random candidates. The random candidates had been evaluated directly on validation, so this single-point diagnostic is descriptive (Table B24).

### Parameter ranges and solution selection

A matched range diagnostic compared the supplied knowledge-infrastructure bounds with bounds widened threefold before transformation and non-negativity constraints, with two planned arms per family at a nominal budget of 150 and seed 0 (Fig. A3a; Table B22). Eight arms yielded assessable validation results. The WOFOST supplied-range sensitivity screen returned no finite losses and its widened-range arm completed eighty optimizer evaluations; the other completed searches recorded 160 optimizer evaluations; the widened MODFLOW 6 search completed, but a split probe left validation inconclusive because the metric response was within 2.00% on both splits. Cached selection diagnostics compared the report-selected member with the existing candidate minimizing the mean of max(0, 1 − NSE) across targets, both assessed on calibration records for all twelve targets of the six-family application (Fig. A3c; Table B23); crop NSE and bias are diagnostic quantities distinct from the optimized NRMSE. Discharge protection was assessed with two selection rules on the same forty-member CRHM front after 120 evaluations:

protection changed validation discharge NSE from −0.16 to 0.34 and snow NSE from −2.08 to 0.41 and increased accepted components from zero to two of four, while both selections remained unaccepted and discharge performance remained below the initial baseline (Fig. A3b). In WOFOST, both calibration NRMSE values and one validation value met the 30.00% band, yet neither validation component satisfied joint acceptance because both worsened relative to the baseline (Fig. A3d).

### Sampling adapter

The DREAM adapter maximizes negative loss, as does DDS, whereas SCE-UA minimizes loss. A HYMOD sampling demonstration at a nominal budget of 3,000 and seed 0 produced seven chains, 34 filled iterations and 119 post-burn-in states. The demonstration shows that the adapter runs; it contributes no calibration endpoint, and convergence, effective sample size, likelihood adequacy and uncertainty coverage were not evaluated.